\documentclass[journal]{IEEEtran}

\usepackage{amsmath,amssymb,mathtools,amsthm}
\usepackage{graphicx}
\usepackage{booktabs}
\usepackage{multirow}
\usepackage{colortbl}
\usepackage[table]{xcolor}
\usepackage{rotating}
\usepackage{array}
\usepackage{float}
\usepackage{pifont}
\usepackage{enumitem}
\usepackage{verbatim}
\usepackage[hidelinks]{hyperref}
\usepackage{tabularx}
\usepackage{array}

\title{Fly0: Persistent Metric Anchoring for Zero-Shot Aerial Vision-Language Navigation}

\author{Zhenxing Xu, Yihong Lu, Weidong Bao, Zhengqiu Zhu,
Jingxuan Zhou, Zhichuang Wang, Ji Wang, Lihua Liu, and
Wei He,~\IEEEmembership{Fellow, IEEE}%
\thanks{Zhenxing Xu, Weidong Bao, Zhengqiu Zhu, Jingxuan Zhou,
Ji Wang, and Lihua Liu are with the College of Systems Engineering,
National University of Defense Technology.}%
\thanks{Yihong Lu is with Phytium Technology.}%
\thanks{Zhichuang Wang and Wei He are with the University of Science
and Technology Beijing.}%
}

\begin{document}
\maketitle

\begin{abstract}
Language-guided UAVs must interpret open-vocabulary goals while producing collision-free motion in continuous three-dimensional space. A central obstacle is that language and flight control operate on different representations and time scales: an MLLM can identify a semantic referent, but a UAV controller requires a stable metric objective that remains valid under occlusion, viewpoint change, and communication delay. Here we formulate zero-shot aerial vision-language navigation as the problem of \emph{persistent metric anchoring}: estimating, validating, and maintaining a relation-conditioned 3D navigation goal from transient semantic evidence. We instantiate this formulation in Fly0, where the MLLM is constrained to a structured grounding contract and never acts as the recurrent flight controller. The interface lifts grounded image regions to uncertain world-frame candidates, converts object referents into reachable free-space anchors according to spatial relations, rejects inconsistent semantic updates, and reuses validated anchors between low-frequency semantic queries. This separation allows semantic reasoning to run at 0.5--1 Hz while LiDAR-based planning and PX4 control remain at 50--100 Hz. We evaluate Fly0 on AerialVLN, OpenFly, and Fly0-Real, a real-world benchmark with 18 scenes and 800 executed flight episodes. Fly0 improves success rate over the strongest zero-shot baseline by 23.7 percentage points in simulation and 21.4 points in real flight, with lower 3D localization error, collision rate, and tail latency. Matched-backbone and matched-planner controls show that these gains arise from the goal interface rather than from the planner alone. The results suggest that reliable UAV vision-language navigation depends not only on stronger multimodal models, but also on interfaces that preserve uncertainty, relation semantics, and metric persistence across the semantic-control boundary. Our repository includes prompts, parsers, evaluation scripts, baseline adapters, and an auditable subset of Fly0-Real at \href{https://anonymous.4open.science/r/Fly0-BE71/}{https://anonymous.4open.science/r/Fly0-BE71/}.
\end{abstract}

\begin{IEEEkeywords}
Aerial navigation, vision-language navigation, multimodal large language model, geometric planning, zero-shot navigation.
\end{IEEEkeywords}

\section{Introduction}
Vision-language navigation (VLN) asks an embodied agent to follow open-ended language instructions in an unseen environment. For UAVs, this problem is particularly challenging because the agent must combine semantic understanding with safety-critical motion generation in continuous 3D space \cite{1}. Recent aerial VLN studies have introduced increasingly realistic simulators, real-world datasets, and VLM/LLM-based navigation agents \cite{5,4,6,airnav2026,vlfly2025,openvln2025,lookasidevln2026,aerial_vln_survey2026}. In parallel, the automation and UAV-control literature has emphasized that visual navigation must respect onboard state estimation, visibility constraints, limited field of view, and collision-free control \cite{zheng2015jas,liu2021nanoquad,yang2024rmpc,peng2025multi}. These two research threads point to the same requirement: language grounding cannot be separated from metric state, visibility, and safety constraints. Unlike ground robots operating at relatively low speeds, aerial platforms must respond to obstacles, drift, and limited field of view within tight control budgets, which makes the interface between language reasoning and geometric control a first-class systems problem rather than a detail of implementation.

Existing UAV-VLN approaches largely fall into two categories. End-to-end or supervised policies \cite{1,3,4,5,6,unified_aerial_vln2025,openvln2025} directly map observations to actions, but their black-box structure often leads to poor interpretability and brittle sim-to-real transfer. MLLM-based and zero-shot agents \cite{8,9,10,11,12,vlfly2025,visa_aerial_vln2026,lookasidevln2026} improve open-vocabulary semantic reasoning, yet many of them still use the model inside the recurrent control loop. In practice, this design asks the MLLM to solve two mismatched problems at once: low-frequency semantic disambiguation and high-frequency geometric control.

\begin{figure}[t]
  \centering
  \includegraphics[width=0.5\textwidth]{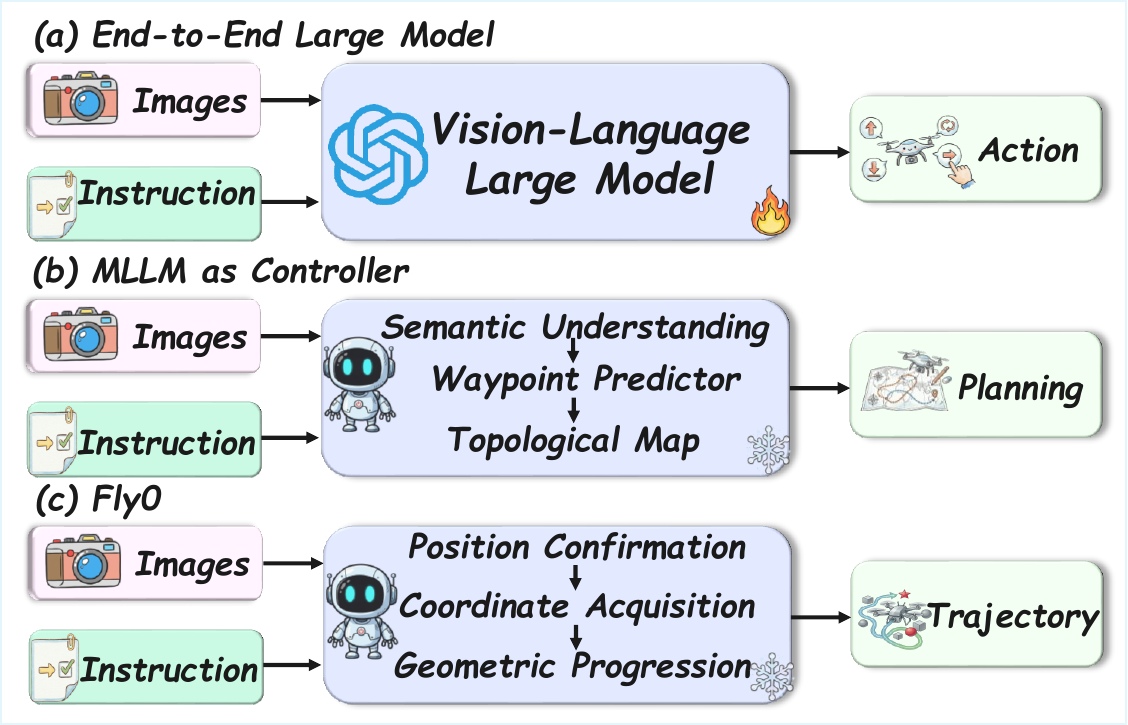}
  \caption{Three design patterns for UAV vision-language navigation. (a) End-to-end policies map observations directly to actions. (b) Controller-style MLLM agents repeatedly invoke a large model inside the decision loop. (c) Fly0 invokes the MLLM at low frequency to estimate a persistent goal anchor and delegates high-rate motion generation to a geometric planner.}
  \label{fig0}
\end{figure}

This coupling creates three recurrent failure modes, illustrated in Fig. \ref{fig0}. First, there is a \textbf{granularity mismatch}: language describes sparse semantic goals, whereas flight control requires dense metric signals. Second, there is a \textbf{latency mismatch}: auto-regressive multimodal reasoning is too expensive to run at the same rate as onboard planning and control. Third, there is a \textbf{memory mismatch}: if the target is temporarily occluded or leaves the image plane, controller-style MLLM agents often lose the goal because they lack a persistent metric anchor in the world frame. This concern is consistent with recent navigation-memory studies showing that first-person policies benefit from explicit spatial memory rather than relying only on the current observation \cite{cognitive_navigation_jas2024}.

Motivated by this observation, we reformulate aerial VLN as the construction of a \emph{semantic-to-geometric interface}. The key question is not whether a planner should be used, but how an open-vocabulary language target should be converted into a persistent, uncertainty-aware, and navigable metric goal. Based on this view, Fly0 treats the MLLM strictly as a semantic localizer and delegates motion execution to a geometric subsystem. Specifically, Fly0 first predicts a structured semantic grounding output consisting of a target point, target region, spatial relation token, and confidence score. It then lifts the grounded region into 3D using robust depth aggregation, estimates uncertainty at the semantic-geometric bridge, and converts the object-centric estimate into a relation-aware navigable goal. Finally, this goal is stored as a persistent 3D anchor that can be reused between low-frequency semantic updates while a LiDAR-based planner runs onboard at high frequency.

This design differs from a simple ``MLLM + planner'' cascade in three important ways. First, the output of the language model is not treated as a transient image coordinate, but as the input to a world-frame goal memory that remains actionable even after the target leaves the image plane. Second, the 2D-to-3D lifting step is uncertainty-aware: it aggregates depth over a grounded region, rejects unreliable projections, and triggers re-grounding when ambiguity becomes large. Third, the interface is relation-aware: commands such as ``fly to the front of the tree'' or ``hover above the bushes'' are mapped to reachable free-space anchors rather than raw object-surface points.

Fly0 should therefore be read as an interface contribution rather than as a claim that every component is new in isolation. Region grounding, spatial memory, semantic mapping, and geometric planning all have clear precedents in modular navigation \cite{chen2021topological,georgakis2022cross,huang2022visual,cognitive_navigation_jas2024,ego}. What remains under-specified for UAV-VLN is the \emph{contract} between uncertain open-vocabulary semantic observations and a high-rate safety-critical planner. Fly0 defines this contract through a structured semantic output with explicit uncertainty and relation tokens, a world-frame anchor update rule that can reject or defer semantic observations before they contaminate the control state, and a free-space goal synthesis stage that makes the anchor directly consumable by a UAV planner. The emphasis of this paper is therefore on interface design and evidence under matched sensing, model, and planner conditions, not on introducing a new low-level planner or a new general-purpose VLM. The main contributions of this paper are threefold:
\begin{enumerate}[leftmargin=*]
\item We formulate zero-shot UAV-VLN as a semantic-to-geometric interface problem and propose a persistent metric goal representation that decouples low-frequency MLLM reasoning from high-frequency geometric control.
\item We introduce a structured grounding and goal-formation pipeline, including region-based semantic grounding, uncertainty-aware 3D lifting, relation-aware navigable goal generation, and persistent goal memory with adaptive re-grounding.
\item We provide a systematic evaluation protocol with fairness-controlled baselines, independent 2D/3D grounding analysis, safety and latency measurements, network-robustness tests, and a real-world UAV benchmark demonstrating gains beyond planner replacement alone.
\end{enumerate}
\section{Related Work}
\subsection{Policy Learning for Vision-Language Navigation}

One influential line of VLN research treats instruction following as a policy-learning problem. In early indoor settings, language embeddings and visual observations were encoded jointly and decoded into navigation actions \cite{anderson2018vision,fried2018speaker}. Later models replaced recurrent encoders with stronger attention mechanisms \cite{hao2020towards,hong2021vln}, introduced reinforcement-learning objectives for cross-modal alignment \cite{wang2019reinforced}, and extended the same policy-centric view to aerial benchmarks and UAV datasets \cite{1,3,4,5,6,unified_aerial_vln2025,openvln2025}.

This formulation is attractive because perception, language interpretation, and motion choice can be optimized in a single training loop. Its cost is that the learned policy often hides the intermediate commitments that matter in flight: which object was grounded, how the target was localized in 3D, and whether a proposed motion is dynamically and geometrically feasible. When a UAV enters a new outdoor scene, changes viewpoint quickly, or observes repeated objects, such entanglement makes errors difficult to diagnose and can weaken transfer beyond the training distribution \cite{zhu2020vision}. Fly0 therefore does not use a learned policy to emit control commands directly; it exposes the grounding and metric-goal construction steps as explicit system states.

\subsection{MLLM Agents for Zero-Shot Navigation}
MLLMs have made zero-shot and few-shot navigation agents possible by allowing instructions, images, and intermediate reasoning to be handled through prompts rather than task-specific training \cite{zhou2024navgpt,huang2022visual}. Existing agents commonly ask the model to answer navigation questions, choose from candidate actions, generate code-like plans, or output waypoint descriptions that are executed by a robot interface \cite{liang2022code,shah2023lm}. Recent aerial systems further exploit large-model reasoning for open-vocabulary target selection, progress assessment, and onboard decision making \cite{8,9,10,11,12,vlfly2025,visa_aerial_vln2026,lookasidevln2026,onfly2026}.

Although these methods greatly broaden the semantic vocabulary of AVLN, a prompt response is not automatically a stable metric objective. If the model is queried inside the fast decision loop, inference delay and view-to-view variability can appear as pauses, oscillatory targets, or inconsistent stopping behavior. If the model is used only to produce a coarse textual waypoint, the geometric controller receives little information about uncertainty or spatial relations. Fly0 takes a narrower role for the MLLM: the model supplies structured semantic evidence at low frequency, while anchor validation, free-space goal construction, and trajectory execution are handled by deterministic geometric modules on the UAV side.

\subsection{Spatial Representations and Geometric Backends}

Modular navigation systems separate semantic perception from localization, mapping, and planning. Topological-memory methods store places and connectivity \cite{chen2021topological}, while metric semantic maps attach object or language information to a geometric frame \cite{georgakis2022cross}. These representations are useful when the task requires global recall or repeated visits, but maintaining a dense semantic map can be expensive for agile UAVs because pose drift, narrow fields of view, and rapidly changing perspectives continuously affect the map state.

Fly0 uses a smaller spatial state. Instead of accumulating a scene-scale semantic map, it keeps a validated task anchor that is sufficient for the active instruction clause. The design still benefits from prior work on semantic memory, visual grounding, and geometry-aware navigation \cite{huang2022visual,cognitive_navigation_jas2024}, but it avoids making the semantic map itself the main planning substrate. The local obstacle representation and trajectory optimizer remain geometric, which makes collision checking and dynamic feasibility independent of the MLLM response format.

\subsection{Positioning Relative to Existing Modular Systems}
Fly0 is best understood as an interface layer between open-vocabulary grounding and a safety-critical UAV planner. Prior modular pipelines have already explored several ingredients: grounding a target from a view, projecting image evidence into 3D, storing spatial memory, and planning collision-free motion. The distinction here is the contract that joins these ingredients. Fly0 requires the semantic branch to return a point, region, relation token, and confidence; converts the grounded evidence into a world-frame candidate with uncertainty checks; transforms object-centric references into reachable free-space anchors; and updates the planner-facing goal only when the new evidence is consistent with the existing state.

This emphasis differs from systems that pass a transient image point directly to a planner, from object-navigation methods that stop after 3D localization, and from semantic-map approaches that preserve a broad scene representation even when the current task needs only one reliable endpoint. It also differs from language-front-end obstacle-avoidance systems in which language selects a coarse destination but the semantic-to-geometric handoff is left implicit. The problem addressed by Fly0 is precisely this handoff: how to make uncertain MLLM grounding usable by a high-rate UAV controller without treating either component as a substitute for the other.
\begin{figure*}[t]
  \centering
  \includegraphics[width=\textwidth]{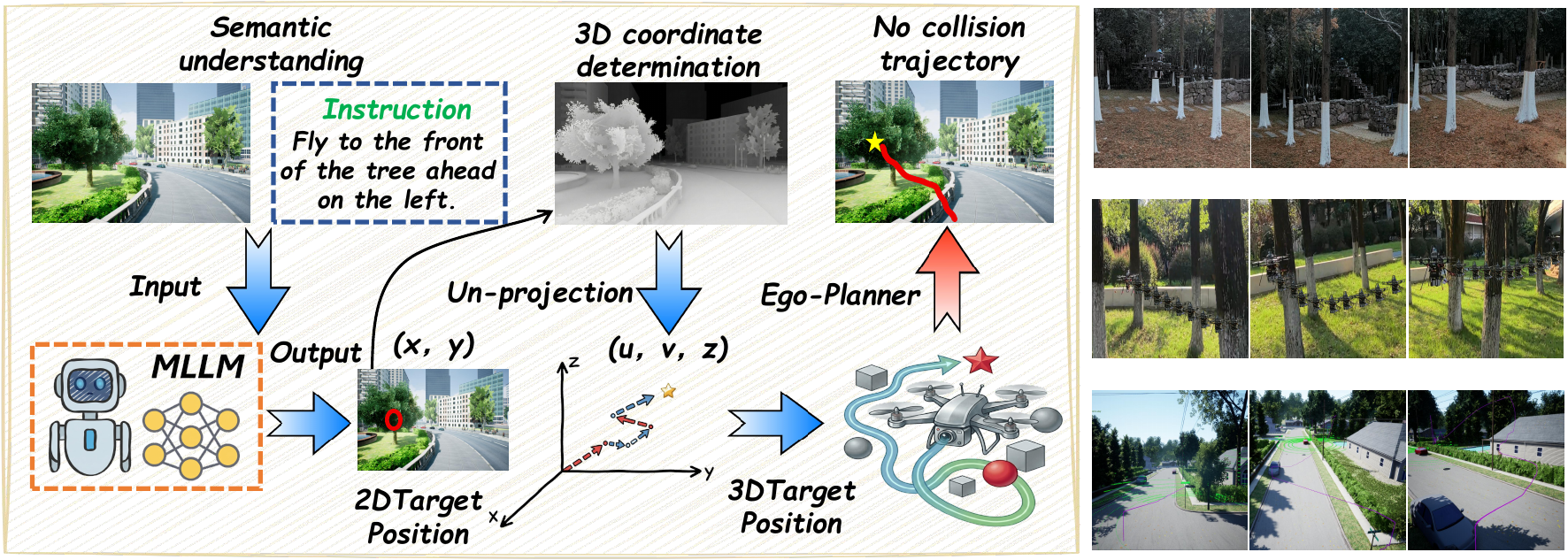}
  \caption{System overview of Fly0. The cloud module performs low-frequency semantic grounding and returns a structured output containing a point, region, relation token, and confidence. The edge module performs uncertainty-aware 3D lifting, relation-aware goal generation, persistent anchor update, and high-frequency LiDAR-based trajectory optimization using Ego-Planner \cite{ego}. The right panel illustrates the hardware and simulation experiments conducted with Fly0.}
  \label{fig1}
\end{figure*}
Fig. \ref{fig1} expands the decoupled pattern in Fig. \ref{fig0}(c) into the concrete Fly0 dataflow: semantic grounding is executed as a low-frequency cloud-side service, while anchor validation, free-space goal synthesis, and collision-aware replanning remain on the UAV-side stack.
\section{Problem Formulation}
Let the UAV state at time $t$ be $\mathbf{x}_t$, and let the multimodal observation be $o_t=\{I_t, D_t, \mathcal{P}^{lidar}_t, T_t\}$, consisting of the RGB image, aligned depth map, LiDAR point cloud, and vehicle pose. Classical VLN seeks a policy $\pi$ that directly maps $(o_t, L)$ to a control action $a_t$, such that
\begin{equation}
\mathbf{x}_{t+1} = \mathcal{T}(\mathbf{x}_t, a_t), \qquad a_t = \pi(o_t, L).
\end{equation}
However, for UAVs, directly learning or prompting $\pi$ conflates semantic disambiguation with high-frequency control.

We instead factor the problem into a semantic-to-geometric interface $\mathcal{I}$ and a geometric controller $\Pi$. The interface predicts a persistent navigable goal anchor $\bar{\mathbf{g}}_t \in \mathbb{R}^3$ from the current observation, instruction, and previous anchor:
\begin{equation}
\bar{\mathbf{g}}_t = \mathcal{I}(I_t, D_t, T_t, L, \bar{\mathbf{g}}_{t-1}).
\end{equation}
The controller then computes high-frequency actions from the local map and anchor:
\begin{equation}
a_t = \Pi(\mathbf{x}_t, \mathcal{M}_t, \bar{\mathbf{g}}_t),
\end{equation}
where $\mathcal{M}_t$ is the local obstacle map built from LiDAR. A navigation episode succeeds if the terminal state falls within a tolerance $\delta$ of the ground-truth target and no collision occurs:
\begin{equation}
\mathcal{S}_{rate} = \mathbb{P}\left(\|\mathbf{x}_{T} - \mathbf{x}_{goal}\|_2 < \delta \;\wedge\; \text{safe}(\tau)\right).
\end{equation}
Here $T$ is the terminal time of an episode, $\mathbf{x}_{T}$ is the final UAV position, $\mathbf{x}_{goal}$ is the annotated target endpoint, $\delta$ is the success tolerance, and $\tau=\{\mathbf{x}_t\}_{t=0}^{T}$ denotes the executed trajectory. The predicate $\text{safe}(\tau)$ is true only when the trajectory contains no collision, safety-pilot intervention, or clearance violation under the benchmark protocol; $\mathbb{P}(\cdot)$ denotes the empirical probability over matched instruction-start episodes.
Under this formulation, the central challenge of UAV-VLN is to design $\mathcal{I}$ so that open-vocabulary language goals become reliable metric anchors for the geometric controller.

\section{Method}

\begin{figure*}[t]
  \centering
  \includegraphics[width=\textwidth]{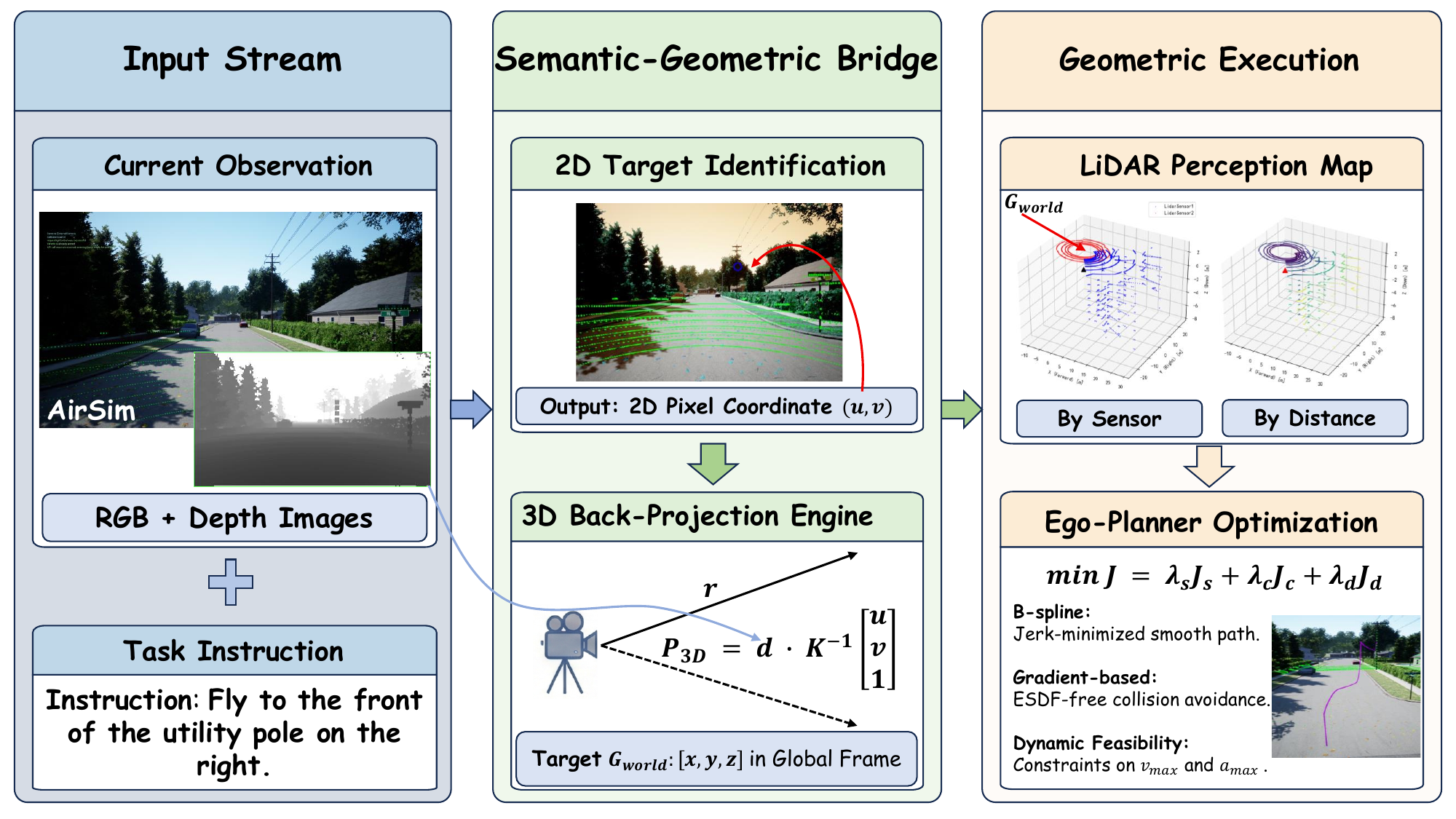}
  \caption{Execution pipeline of Fly0. The semantic branch predicts a structured target description, while the geometric branch performs uncertainty-aware 3D lifting, free-space projection, and LiDAR-based replanning.}
  \label{fig1.5}
\end{figure*}

Fly0 is built around a semantic-to-geometric interface that converts sparse language grounding into a persistent, navigable, and uncertainty-aware metric target. The architecture contains five stages: structured semantic grounding, uncertainty-aware 3D goal lifting, relation-aware navigable goal generation, persistent goal memory, and LiDAR-based trajectory optimization. Fig. \ref{fig1.5} shows the runtime pipeline and clarifies where semantic evidence is converted into the metric anchor consumed by the planner.

\subsection{Referring Semantic Grounding with Structured Outputs}

Let $\Phi_{VLM}$ denote the multimodal language model used for semantic grounding. At update time $t_s$, the UAV sends the current RGB image $I_{t_s}$ and instruction $L$ to $\Phi_{VLM}$ and requests a structured JSON output rather than free-form actions:
\begin{equation}
q_{t_s} = \Phi_{VLM}(I_{t_s}, L) = \{\mathbf{p}_{t_s}, \mathbf{b}_{t_s}, r_{t_s}, c_{t_s}\},
\end{equation}
where $\mathbf{p}_{t_s}=(x,y)$ is the grounded point, $\mathbf{b}_{t_s}=(x_1,y_1,x_2,y_2)$ is a grounded region, $r_{t_s}$ is a spatial-relation token (e.g., \texttt{front}, \texttt{left}, \texttt{above}, \texttt{near}), and $c_{t_s} \in [0,1]$ is the model-reported confidence. The prompt used to obtain this output is given in Appendix \ref{apd-prompt}.

Across all benchmarks and models, we use the same semantic schema with four required fields: \texttt{point}, \texttt{box}, \texttt{relation}, and \texttt{confidence}. Provider-specific wrappers are used only to enforce JSON mode when available; field names, relation vocabulary, and downstream parsing are fixed. This structured output is important for two reasons. First, the region $\mathbf{b}_{t_s}$ provides spatial support for robust depth aggregation, making the system less sensitive to noisy single-pixel measurements. Second, the relation token exposes the semantic offset requested by the instruction so that the downstream module can convert an object-centric estimate into a navigable free-space goal. If the JSON parser detects malformed coordinates, degenerate boxes, unsupported relation tokens, or confidence values outside $[0,1]$, the semantic update is rejected and the system retains the previous valid anchor. 

For multi-step instructions, a deterministic clause parser decomposes the command into an ordered list of clause-level sub-goals $\{L^{(m)}\}_{m=1}^{M}$ before any MLLM call. The parser splits temporal connectives such as \emph{then}, \emph{after}, \emph{before}, and comma-separated imperative clauses, while preserving spatial relation phrases inside each clause. If no unambiguous split is found, the instruction is treated as a single clause. Only the active clause $L^{(m_t)}$ is grounded at runtime with the same JSON schema, and the system advances to $m_t+1$ once the vehicle enters the completion radius of the current clause or the clause-specific target is not visible for three consecutive semantic updates. Parser errors are not corrected using ground truth; they propagate through the normal end-to-end evaluation. This mechanism is illustrated in Appendix \ref{apd-trajectories} and prevents a single persistent anchor from conflating sequential targets.

\subsection{Uncertainty-aware 3D Goal Lifting}

Single-point back-projection is fragile for aerial navigation because the grounded pixel may fall on background clutter, object boundaries, sky, or missing depth values. We therefore aggregate depth over a local support region. Around each grounded target, we define a valid depth set
\begin{equation}
\Omega_{t_s} = \Omega(\mathbf{p}_{t_s}, \mathbf{b}_{t_s}), \qquad
\mathcal{D}_{t_s} = D_{t_s}(\Omega_{t_s}) \cap (d_{min}, d_{max}),
\end{equation}
where $\Omega(\mathbf{p}_{t_s}, \mathbf{b}_{t_s})$ is the intersection of the predicted box and a square patch centered at $\mathbf{p}_{t_s}$. For the 896$\times$896 grounding input used in all experiments, the patch side length is adaptive:
\begin{equation}
s = 2\left\lfloor\frac{\mathrm{clip}(0.35\min(w_b,h_b),21,96)}{2}\right\rfloor + 1,
\end{equation}
where $w_b$ and $h_b$ are the box width and height in pixels. Thus the support is at least $21\times21$ pixels, at most $97\times97$ pixels, and remains odd-sized so that it is centered on $\mathbf{p}_{t_s}$. We use $d_{min}=0.8$ m and $d_{max}=35$ m for the ZED 2i and simulated depth cameras. Let $\eta_{t_s}=|\mathcal{D}_{t_s}|/|\Omega_{t_s}|$ denote the valid-depth ratio. We estimate depth using the median rather than a single sample:
\begin{equation}
\hat d_{t_s} = \mathrm{median}(\mathcal{D}_{t_s}), \qquad \sigma^2_{d,t_s} = \mathrm{Var}(\mathcal{D}_{t_s}).
\end{equation}
The variance is computed after removing the lowest and highest 10\% of valid depths to reduce the influence of depth discontinuities at object boundaries. The pair $(\hat d_{t_s}, \sigma^2_{d,t_s})$ gives both the 3D lifting estimate and a local uncertainty signal. If $\eta_{t_s}<\tau_{occ}=0.35$ or $\sigma^2_{d,t_s}>\tau_{var}=2.25$ m$^2$, the semantic update is marked unreliable and a re-grounding request is triggered.

Using the box center $\tilde{\mathbf{p}}_{t_s}$ and the estimated depth $\hat d_{t_s}$, we compute the object-centric 3D target in the camera frame:
\begin{equation}
\mathbf{g}^{obj}_{cam} = \hat d_{t_s} K^{-1}
\begin{bmatrix}
\tilde x_{t_s} \\ \tilde y_{t_s} \\ 1
\end{bmatrix}.
\end{equation}
It is then transformed to the world frame through the camera-to-body extrinsic $T_{IC}=\{R_{IC}, t_{IC}\}$ and the body-to-world pose $T_t=\{R_t,t_t\}$:
\begin{equation}
\mathbf{g}^{obj}_{world} = R_t(R_{IC}\mathbf{g}^{obj}_{cam} + t_{IC}) + t_t.
\end{equation}
We define the semantic-geometric confidence of this estimate as
\begin{equation}
w_{t_s} = c_{t_s}\exp\left(-\sigma^2_{d,t_s}/\tau^2_{\sigma}\right),
\end{equation}
where $\tau_{\sigma}=1.5$ m. Candidates with $c_{t_s}<0.35$ or $w_{t_s}<0.25$ are rejected before anchor fusion; the remaining candidates use $w_{t_s}$ to update the persistent goal memory. All interface constants are fixed before evaluation and summarized in Table \ref{tab:impl_constants} of Appendix \ref{apd-implementation}.

\subsection{Relation-aware Navigable Goal Generation}

The grounded object location is not always the desired navigation endpoint. Instructions such as ``fly to the front of the tree'' or ``hover above the bushes'' refer to reachable free-space locations near the object, not to the object surface itself. To address this gap, we convert the object-centric target into a navigable goal:
\begin{equation}
\mathbf{g}^{nav}_{t_s} = \Pi_{\mathcal{F}_t}\left(\mathbf{g}^{obj}_{world} + \Delta(r_{t_s}, \psi_{t_s}, \mathcal{R}_{t_s})\right),
\end{equation}
where $\psi_{t_s}$ is the UAV heading at the time of grounding, $\mathcal{R}_{t_s}$ denotes the local relation frame frozen at the same instant, $\Delta(\cdot)$ is a relation-aware offset, and $\Pi_{\mathcal{F}_t}$ projects the raw target to the closest collision-free point in the local free-space set $\mathcal{F}_t$.

In our implementation, the offset is parameterized by a small set of interpretable rules. Let $\mathbf{e}_f=[\cos\psi_{t_s},\sin\psi_{t_s},0]^\top$ be the frozen heading axis, $\mathbf{e}_l=[-\sin\psi_{t_s},\cos\psi_{t_s},0]^\top$ the lateral axis, $\mathbf{e}_z=[0,0,1]^\top$ the world vertical axis, and $\mathbf{e}_{vis}$ the horizontal unit vector from the grounded object to the UAV at grounding time. The important convention is that relation tokens are \emph{observation-conditioned} rather than object-canonical: \texttt{front} uses the visible side of the object, i.e., $+2.0\mathbf{e}_{vis}$ m; \texttt{behind} uses $-2.0\mathbf{e}_{vis}$ m; \texttt{left} and \texttt{right} use $\pm2.0\mathbf{e}_l$ m; \texttt{above} and \texttt{below} use $+1.5\mathbf{e}_z$ m and $-1.0\mathbf{e}_z$ m; and \texttt{near} uses a shorter $+1.5\mathbf{e}_{vis}$ m standoff. \texttt{around} is implemented by sampling eight candidates on a 2.5 m horizontal ring and selecting the closest free candidate to the incoming direction. \texttt{inside} uses the projected centroid of the grounded region. \texttt{between} is handled by grounding the two noun phrases in the clause and projecting the midpoint of their 3D estimates. Freezing $\mathcal{R}_{t_s}$ prevents the target from drifting when the vehicle later changes heading. The full relation table is provided in Table \ref{tab:relation_offsets} of Appendix \ref{apd-implementation}.

The free-space set $\mathcal{F}_t$ is built from the same LiDAR ring-buffer occupancy map used by Ego-Planner, not from an offline map or from the MLLM. The ring buffer covers a 30 m $\times$ 30 m $\times$ 8 m local window at 0.2 m voxel resolution. LiDAR ray casting marks traversed voxels as free and endpoints as occupied; occupied voxels are inflated by 0.45 m, and unknown voxels are not considered valid goal locations. $\Pi_{\mathcal{F}_t}$ searches a 0.25 m lattice around the raw relation target with an expanding radius up to 6 m, keeps candidates inside the altitude band $[0.8,8.0]$ m and outside inflated obstacles, and returns the candidate minimizing distance to the raw target plus a small smoothness term to the current anchor. If no candidate passes the Ego-Planner collision check, the semantic update is rejected and the previous anchor is retained.

\subsection{Persistent Goal Memory and Low-Frequency Re-grounding}

To keep the target actionable when it becomes temporarily invisible, Fly0 maintains a persistent 3D goal anchor $\bar{\mathbf{g}}_t$. When a new valid semantic update arrives, the anchor is refreshed by confidence-weighted fusion:
\begin{equation}
\alpha_{t_s} = \mathrm{clip}(w_{t_s}, \alpha_{min}, \alpha_{max}).
\end{equation}
\begin{equation}
\bar{\mathbf{g}}_{t_s} =
\alpha_{t_s}\mathbf{g}^{nav}_{t_s} +
(1-\alpha_{t_s})\bar{\mathbf{g}}_{t_s^-}.
\end{equation}
Between semantic updates, the planner continues to track $\bar{\mathbf{g}}_t$ without invoking the MLLM. We use $\alpha_{min}=0.20$ and $\alpha_{max}=0.80$ after the confidence and depth validity checks described above. By default, semantic re-grounding runs at 0.5 Hz. The update frequency increases to 1.0 Hz when the target is within 12 m, when the depth uncertainty rises above threshold, or when repeated-object ambiguity is detected.

Persistent anchors can also amplify an early false positive if they are updated too aggressively. To reduce this risk, Fly0 performs a consistency check before committing a new anchor. Let
\begin{equation}
\rho_{t_s} = \|\mathbf{g}^{nav}_{t_s} - \bar{\mathbf{g}}_{t_s^-}\|_2
\end{equation}
denote the anchor-jump residual. We set $\tau_{jump}=3.0$ m and $\tau_{agree}=1.0$ m. A confidence value is considered close to the decision boundary if $|c_{t_s}-0.55|\leq0.10$. Repeated-object ambiguity is detected automatically from the active clause and recent semantic observations: the parser extracts the head noun phrase, and a rolling 10 s cache is keyed by this phrase; if two valid or tentative 3D candidates for the same phrase are separated by more than 1.5 m, or their image boxes have IoU below 0.3 while both have $c_{t_s}>0.45$, the phrase is marked repeated until the active clause is completed. If $\rho_{t_s}>\tau_{jump}$ while either repeated-object ambiguity is active or the confidence is close to the boundary, the update is marked \emph{tentative} rather than immediately fused. Tentative updates require either two consecutive consistent observations within $\tau_{agree}$ meters or a second viewpoint obtained after a short safety-constrained motion before the anchor is replaced. If neither confirmation condition is met, the previous anchor is retained and the system requests re-grounding.

If the cloud-side semantic service becomes temporarily unavailable, the UAV keeps following the last valid anchor for up to $T_{hold}=4$ s. If no valid goal is recovered within this horizon, the vehicle switches to hover mode; after $T_{fail}=10$ s, it triggers the pre-defined return-to-home or assisted-landing routine. This fallback policy is essential for safe outdoor deployment and is evaluated in Sec. V-D.

\subsection{LiDAR-based Safe Trajectory Optimization}
After the anchor module accepts a goal $\bar{\mathbf{g}}_t$, Fly0 switches to a purely geometric execution path. The MLLM is not queried by the motion controller; it only changes the planner input when a later semantic update passes the validation rules described above. We use Ego-Planner \cite{ego} as the local optimization backend and keep the planner contribution separate from the semantic-to-geometric interface proposed in this paper.

\subsubsection{Local Obstacle Buffer}
At each planning cycle, the onboard LiDAR returns a point set $\mathcal{P}_{t}^{lidar}$ in the sensor frame. These points are expressed in the world frame before they are inserted into the local collision buffer:
\begin{equation}
\mathbf{p}_{world} = R_{t} \cdot (R_{ext} \cdot \mathbf{p}_{l} + \mathbf{t}_{ext}) + \mathbf{t}_{t},
\end{equation}
where $\mathbf{p}_{l} \in \mathcal{P}_{t}^{lidar}$, $R_{ext}$ and $\mathbf{t}_{ext}$ are the fixed LiDAR-to-body calibration, and $R_t,\mathbf{t}_t$ are the current body pose estimated onboard. The transformed points $\mathcal{P}_{t}^{world}$ update the same sliding-window ring-buffer map used by the goal-projection stage, so the free-space check in Sec. IV-C and the downstream trajectory optimizer query a consistent obstacle representation.

\subsubsection{Planner Interface and Objective}
The planner receives the current state, the local obstacle buffer, and the accepted anchor $\bar{\mathbf{g}}_t$. Following Ego-Planner \cite{ego}, the executable path is represented as a uniform B-spline with control points $\{\mathbf{p}_i\}_{i=0}^{n}$:
\begin{equation}
\Phi(s) = \sum_{i=0}^{n} \mathbf{p}_i N_{i,k}(s), \qquad s \in [0,n-k+1],
\end{equation}
where $N_{i,k}$ is the order-$k$ basis function. This parameterization lets the optimizer adjust a compact control polygon while still producing a smooth continuous trajectory for PX4 tracking.

For completeness, we summarize the cost terms used in our integration. The optimized control polygon minimizes
\begin{equation}
\min_{\{\mathbf{p}_i\}} \mathcal{J}(\mathbf{p}) =
\lambda_s \mathcal{J}_s + \lambda_c \mathcal{J}_c + \lambda_d \mathcal{J}_d,
\end{equation}
with $\lambda_s=1.0$, $\lambda_c=10.0$, and $\lambda_d=1.0$ in all experiments. The first term discourages abrupt changes in the third finite difference of the control points,

\begin{equation}
\mathcal{J}_s = \sum_{i=1}^{n-2} \|\Delta^3 \mathbf{p}_i\|_2^2,
\end{equation}
which reduces jerk and keeps the onboard view stable during flight. The obstacle term penalizes only those control points whose nearest-obstacle distance is below the clearance margin:

\begin{equation}
\begin{split}
\mathcal{J}_c &= \sum_{i=1}^{n} \rho(d_{safe}-d(\mathbf{p}_i)),\\
\rho(z) &=
\begin{cases} 
0 & \text{if } z \le 0,\\
z^3 & \text{if } z > 0,
\end{cases}
\end{split}
\end{equation}
where $d(\mathbf{p}_i)$ is obtained from the local obstacle buffer and $d_{safe}=0.5$~m. The dynamic-feasibility term softly enforces the velocity and acceleration limits of the airframe:

\begin{equation}
\begin{split}
\mathcal{J}_d = \sum_{i=1}^{n} \Big( & \max(0, \|\mathbf{v}_i\|_2^2 - v_{\text{max}}^2)^2 \\
& + \max(0, \|\mathbf{a}_i\|_2^2 - a_{\text{max}}^2)^2 \Big).
\end{split}
\end{equation}
We use $v_{\text{max}}=4.0$~m/s and $a_{\text{max}}=3.0$~m/s$^2$.

Planning runs in a receding horizon. At 50 Hz, the previous spline is shifted and reused as the initialization for the next solve; replanning is triggered by local-map updates, a newly accepted semantic anchor, or a predicted clearance violation along the active trajectory. PX4 tracks the resulting command stream at 100 Hz, so the control bandwidth is determined by onboard sensing and planning rather than by the lower-frequency semantic service.
\begin{table}[!htbp]
\caption{Statistics of Fly0-Real, the real-world benchmark used for deployment evaluation. Episodes are executed under matched instructions and start poses, and safety-pilot interventions are counted as failures.}
\label{tab:realworld_stats}
\centering
\resizebox{\columnwidth}{!}{%
\begin{tabular}{ll}
\toprule
Item & Value \\
\midrule
Number of scenes & 18 (10 campus, 8 park) \\
Unique instructions / executed episodes & 160 / 800 \\
Instruction authoring & Human-written + manual paraphrase audit \\
Average trajectory length & 48.7 m \\
Object categories & 17 \\
Spatial relation types & 9 \\
Relation-bearing / multi-step instructions & 58\% / 12\% \\
Lighting conditions & Sunny / cloudy / dusk \\
Occlusion-heavy episodes & 31\% \\
Repeated-object episodes & 28\% \\
Dynamic-obstacle episodes & 14\% \\
Physical trials per instruction & 5 \\
Target annotation & Dual annotators + adjudication \\
Human safety interventions & 6.8\% of flights \\
\bottomrule
\end{tabular}}
\end{table}

\section{Experimental Evaluation}
\subsection{Experimental Setup}

\subsubsection{Simulation and Real-World Testbeds}
We evaluate Fly0 in both high-fidelity simulation and physical flight. The simulation stack is built on UE4 and AirSim, which provide realistic vehicle dynamics, controllable weather and lighting, and synchronized RGB-D sensing. For real-world validation, we deploy the system on a custom quadrotor platform detailed in Appendix \ref{appendix a}. Unless otherwise stated, simulation results are averaged over 20 random seeds and real-world results are averaged over 5 physical flights per instruction.

\subsubsection{Datasets and Benchmark Protocol}
We use two public simulation benchmarks and a new real-world benchmark.

\textbf{AerialVLN and OpenFly (Simulation):} These benchmarks contain 24 urban and suburban maps with more than 14,000 navigation instructions. We preserve the official task definitions and termination conditions, and report results on held-out maps without fine-tuning or benchmark-specific prompt adaptation.

\textbf{Fly0-Real (Real World):} We build a held-out deployment benchmark with repeated-object scenes, occlusion, lighting variation, and moving pedestrians or bicycles. Instructions are human-authored and manually paraphrased rather than synthetically templated at evaluation time. For relation-bearing instructions, the benchmark stores a canonical reachable endpoint defined by a pair of annotators using an object reference and a standoff rule; disagreements larger than 1.0 m are resolved by a third annotator. The benchmark is used only for testing rather than learning. Table \ref{tab:realworld_stats} summarizes its statistics.

All methods are evaluated on the same instructions, start poses, stopping rules, and scene splits. In physical flight, RTK-GNSS is used only for target annotation, terminal error measurement, and post-hoc auditing; it is not an input to the online semantic interface, planner, or controller. Onboard execution uses the same PX4/VIO pose estimate for Fly0 and all fairness-controlled variants. A human safety pilot is allowed to intervene if the predicted minimum clearance drops below 0.3 m or if communication is lost. Such interventions are counted as failures and collisions in our statistics.

\subsubsection{Evaluation Metrics}
We organize evaluation as an evidence chain from grounding to control. SR, NE, and SPL follow the standard VLN evaluation protocol \cite{anderson2018vision,fried2018speaker}, pointing and recall-style localization scores follow common referring-grounding practice \cite{kazemzadeh2014referit,yu2016modeling}, and collision, clearance, jerk, and replanning metrics follow UAV planning and constrained visual-control studies \cite{ego,yang2024rmpc}.
\begin{itemize}[leftmargin=*]
\item \textbf{2D grounding metrics:} Pointing Accuracy (20 px threshold), Mean Pixel Error, Recall@50 px, invalid output rate, and ambiguity failure rate.
\item \textbf{3D grounding metrics:} world-frame target localization error, projection error under occlusion, projection error on thin structures, and depth variance rejection rate.
\item \textbf{Navigation metrics:} Success Rate (SR), Navigation Error (NE), Collision Rate (CR), Success weighted by Path Length (SPL), near-miss rate (clearance $<0.5$ m), path-length ratio, jerk, replanning and False-lock.
\item \textbf{System metrics:} mean latency, P95 latency, semantic update frequency, and network degradation robustness.
\end{itemize}
Unless otherwise noted, success requires ending within 5 m of the target with no collision or safety abort. Because this tolerance can be permissive for relation-sensitive episodes, we additionally report a stricter SR@2m for relation-bearing real-world subsets. Simulation results are reported as mean $\pm$ standard deviation over 20 random seeds, and real-world results are averaged over 5 repeated flights per instruction. The pairing unit is the matched instruction-start pair. Continuous metrics (NE, jerk, path ratio, 3D error) use paired $t$-tests on per-pair averages, whereas binary metrics (SR, CR, safety-abort rate) use stratified bootstrap 95\% confidence intervals together with McNemar tests on matched outcomes. For real flights, repeated runs are treated as clustered samples under the same instruction-start pair, and confidence intervals are computed with block bootstrap over instruction identity. Multiple comparisons are controlled separately within each benchmark family using Holm-Bonferroni correction at the 0.05 level.

\subsubsection{Baselines and Fairness Protocol}
We compare against four end-to-end methods (OpenUAV \cite{5}, UAV-Flow \cite{4}, AerialVLN \cite{1}, OpenFly \cite{6}) and four zero-shot agent baselines (NAVGPT \cite{zhou2024navgpt}, STMR \cite{gao2024aerial}, CityNavAgent \cite{8}, SPF \cite{12}). Table \ref{tab:fairness_inputs} summarizes the main sensing, localization, planning, and training assumptions behind these representative baselines. To avoid confounding factors, we additionally construct controlled variants with the same MLLM, the same local planner, and the same sensor stack whenever applicable. Detailed implementation and adaptation rules for each baseline are provided in Appendix \ref{apd-baselines}.

\begin{table}[!htbp]
\caption{Comparison of sensing, localization, planning, and training assumptions of representative baselines. Pose denotes onboard odometry or VIO used at execution time; RTK-GNSS is excluded because it is reserved for annotation and evaluation only. The heterogeneity summarized here motivates the controlled comparisons reported in Table \ref{tab:fairness}.}
\label{tab:fairness_inputs}
\centering
\resizebox{\columnwidth}{!}{%
\begin{tabular}{lccccccc}
\toprule
Method & RGB & Depth & LiDAR & VIO/Pose & Planner & MLLM & Training \\
\midrule
OpenFly & Y & N & N & N & None & Y & Supervised \\
UAV-Flow & Y & N & N & N & None & Y & Supervised \\
NAVGPT & Y & N & N & Y & Heuristic & Y & Zero-shot \\
SPF & Y & Est. & N & Y & Heuristic & Y & Zero-shot \\
Fly0 & Y & Y & Y & Y & Ego & Y & Zero-shot \\
\bottomrule
\end{tabular}}
\end{table}

\begin{table*}[!htbp]
\caption{Independent 2D and 3D grounding evaluation on the AerialVLN-Ground subset and a manually annotated real-world subset (3,200 frames). Point Acc uses a 20 px threshold, and Occl. Err. reports 3D localization error on occlusion-tagged frames.}
\label{tab:grounding}
\centering
\resizebox{\textwidth}{!}{%
\begin{tabular}{l|ccccc|cc}
\toprule
\multirow{2}{*}{Method} & \multicolumn{5}{c|}{2D grounding} & \multicolumn{2}{c}{3D grounding} \\
\cline{2-8}
& Point Acc. $\uparrow$ & Pix. Err. $\downarrow$ & Recall@50 $\uparrow$ & Invalid $\downarrow$ & Ambig. Fail $\downarrow$ & 3D Err. $\downarrow$ & Occl. Err. $\downarrow$ \\
\midrule
SPF-Qwen & 74.6 & 28.3 & 80.7 & 4.1 & 14.7 & 6.3 & 8.6 \\
Fly0 point-only & 81.9 & 21.4 & 88.5 & 2.8 & 10.3 & 4.7 & 7.1 \\
\textbf{Fly0 full} & \textbf{86.4} & \textbf{16.9} & \textbf{92.7} & \textbf{1.2} & \textbf{6.1} & \textbf{3.1} & \textbf{4.8} \\
\bottomrule
\end{tabular}}
\end{table*}

\begin{table*}[!htbp]
\caption{Overall navigation performance on simulation and real-world benchmarks. Results are mean $\pm$ standard deviation over 20 simulation seeds and 5 physical flights per instruction. CR denotes collision rate. Unless otherwise stated, each method uses its native sensing and planning stack.}
\label{tab:overall}
\centering

\setlength{\belowrulesep}{0pt}

\newcommand{\GroupLabel}[1]{%
  \multirow{4}{*}{%
    \rotatebox[origin=c]{90}{%
      \resizebox{0.95cm}{!}{\scriptsize\textbf{#1}}%
    }%
  }%
}

\resizebox{\textwidth}{!}{%
\begin{tabular}{cl|ccc|ccc|ccc}
\toprule

& \multirow{2}{*}{Method}
& \multicolumn{3}{c|}{\cellcolor{gray!15}\textbf{AerialVLN}}
& \multicolumn{3}{c|}{\cellcolor{purple!10}\textbf{OpenFly}}
& \multicolumn{3}{c}{\cellcolor{cyan!10}\textbf{Fly0-Real}} \\

\cline{3-11}

&
& \cellcolor{gray!15}SR $\uparrow$
& \cellcolor{gray!15}NE $\downarrow$
& \cellcolor{gray!15}CR $\downarrow$
& \cellcolor{purple!10}SR $\uparrow$
& \cellcolor{purple!10}NE $\downarrow$
& \cellcolor{purple!10}CR $\downarrow$
& \cellcolor{cyan!10}SR $\uparrow$
& \cellcolor{cyan!10}NE $\downarrow$
& \cellcolor{cyan!10}CR $\downarrow$ \\

\midrule

\GroupLabel{End-to-End}
& OpenUAV
& 20.4$\pm$1.2 & 63.0$\pm$2.8 & 0.41
& 17.8$\pm$1.1 & 74.6$\pm$3.9 & 0.45
& 13.2$\pm$1.6 & 27.4$\pm$2.7 & 0.37 \\

& UAV-Flow
& 36.5$\pm$1.5 & 57.3$\pm$2.5 & 0.33
& 32.1$\pm$1.4 & 69.5$\pm$3.1 & 0.36
& 27.5$\pm$1.9 & 23.6$\pm$2.2 & 0.28 \\

& AerialVLN
& 7.3$\pm$0.6 & 90.2$\pm$4.1 & 0.54
& 6.7$\pm$0.5 & 94.6$\pm$4.8 & 0.58
& 4.0$\pm$0.8 & 49.4$\pm$5.1 & 0.51 \\

& OpenFly
& 37.7$\pm$1.4 & 67.8$\pm$3.6 & 0.29
& 33.6$\pm$1.2 & 64.1$\pm$3.0 & 0.31
& 29.1$\pm$1.7 & 25.6$\pm$2.0 & 0.24 \\

\midrule

\GroupLabel{Zero-Shot}
& NAVGPT
& 34.0$\pm$1.2 & 59.4$\pm$3.1 & 0.23
& 29.3$\pm$1.1 & 61.6$\pm$3.4 & 0.25
& 26.3$\pm$2.0 & 33.9$\pm$3.2 & 0.19 \\

& STMR
& 30.2$\pm$1.1 & 63.8$\pm$3.0 & 0.26
& 27.4$\pm$1.0 & 66.5$\pm$3.2 & 0.29
& 22.6$\pm$1.9 & 37.1$\pm$3.4 & 0.23 \\

& CityNavAgent
& 29.0$\pm$1.0 & 66.3$\pm$3.5 & 0.31
& 25.3$\pm$1.0 & 70.2$\pm$3.8 & 0.34
& 22.2$\pm$1.8 & 38.7$\pm$3.6 & 0.27 \\

& SPF
& \underline{46.7$\pm$1.2} & \underline{39.6$\pm$1.9} & \underline{0.15}
& \underline{42.9$\pm$1.3} & \underline{46.6$\pm$2.4} & \underline{0.17}
& \underline{41.5$\pm$2.1} & \underline{17.8$\pm$1.5} & \underline{0.13} \\

\midrule

& \textbf{Fly0 (ours)}
& \textbf{70.4$\pm$0.9} & \textbf{27.2$\pm$1.1} & \textbf{0.05}
& \textbf{64.7$\pm$1.0} & \textbf{29.5$\pm$1.4} & \textbf{0.06}
& \textbf{62.9$\pm$2.4} & \textbf{13.8$\pm$1.2} & \textbf{0.07} \\

\bottomrule
\end{tabular}}
\end{table*}
For fairness-controlled experiments, we use Qwen2.5VL-32B as the shared MLLM backbone and the same episode budget. Strict interface-isolation experiments additionally fix RGB-D sensing, LiDAR mapping, VIO pose, Ego-Planner, localization noise, and stopping rules, so that only the goal interface differs. Sensor-controlled ablations restrict Fly0 to RGB-only, RGB-D, or RGB-D+LiDAR settings to isolate the contribution of sensing from the interface design.

\subsubsection{Definition of Zero-Shot}
Throughout this paper, \emph{zero-shot} means that Fly0 is not trained, fine-tuned, or adapted on the evaluation scenes, instructions, trajectories, or target annotations from AerialVLN, OpenFly, or Fly0-Real. The same prompt schema, relation vocabulary, geometric constants, and safety fallback rules are used across benchmarks. All reported zero-shot methods therefore operate by inference-time reasoning and planning rather than task-specific policy learning or benchmark-specific adaptation.

\subsubsection{Evaluation Questions and Analysis Scope}
To provide a more comprehensive evaluation of our approach, we formulate six research questions that examine its foundational capabilities, end-to-end navigation performance, fairness of comparison, latency and network robustness, real-world safety, and sensitivity to model scale:
\begin{itemize}[leftmargin=*]
\item \textbf{RQ1 Grounding accuracy:} Does the structured semantic contract improve 2D grounding and 3D target lifting before downstream planning is introduced?
\item \textbf{RQ2 Navigation quality:} Does Fly0 improve success rate, terminal precision, and collision avoidance across AerialVLN, OpenFly, and Fly0-Real?
\item \textbf{RQ3 Fairness controls:} Are the performance gains preserved under matched MLLM backbones, planners, sensors, pose sources, localization noise, and stopping criteria?
\item \textbf{RQ4 Latency and network robustness:} Can low-frequency cloud-based semantic reasoning be integrated with high-frequency onboard mapping, planning, and control under degraded communication?
\item \textbf{RQ5 Real-world generalization and safety:} Does the persistent anchor remain effective under occlusion, repeated objects, dynamic obstacles, and safety-pilot intervention constraints?
\item \textbf{RQ6 Model-scale sensitivity:} How sensitive are grounding and navigation performance to the choice and scale of the MLLM backbone when the Fly0 backend is fixed?
\end{itemize}

\subsection{Results and Analysis}

\subsubsection{Grounding Accuracy}

Table \ref{tab:grounding} shows that the proposed interface improves both semantic grounding and geometric lifting. Compared with a point-only variant, the full Fly0 interface reduces mean pixel error by 4.5 px and 3D localization error by 1.6 m. The gain is larger under ambiguity and occlusion, where region grounding and uncertainty rejection matter most. On thin objects such as poles and branches, the full method reduces projection error from 8.8 m to 6.2 m, indicating that robust depth aggregation is substantially more reliable than single-pixel sampling.

\subsubsection{Overall Navigation Performance}


The overall results in Table \ref{tab:overall} show that Fly0 consistently improves task success, final precision, and safety. On AerialVLN, Fly0 exceeds the strongest zero-shot baseline SPF by 23.7 SR points and reduces NE by 12.4 m. On the real-world benchmark, the gap remains large: Fly0 improves SR by 21.4 points and reduces collision rate from 0.13 to 0.07. The corresponding paired bootstrap confidence interval for the SR gap over SPF is [21.9, 25.5] on AerialVLN and [17.8, 24.6] on Fly0-Real, and the matched binary outcomes are significant under McNemar tests ($p<0.001$ in both cases). These statistics suggest that the observed gain is unlikely to be driven by a few easy scenes.

\subsubsection{Fairness-Controlled Comparisons}

\begin{table*}[t]
\caption{Strict interface-isolation comparison on AerialVLN. The first five rows share the same RGB-D camera, LiDAR map, VIO pose, Qwen2.5VL-32B backbone, Ego-Planner, localization noise, and stopping rule; only the goal interface differs. The final rows show adapted external baselines under the same planner.}
\label{tab:fairness}
\centering
\resizebox{\textwidth}{!}{%
\begin{tabular}{lcccccccl}
\toprule
Method & Shared stack & SR $\uparrow$ & NE $\downarrow$ & CR $\downarrow$ & SPL $\uparrow$ & Jerk $\downarrow$ & False-lock $\downarrow$ & Goal interface \\
\midrule
Pixel point + single depth & Fixed & 62.1 & 31.8 & 0.12 & 0.54 & 2.04 & 0.16 & Transient point goal \\
Box + median depth & Fixed & 64.0 & 30.7 & 0.11 & 0.56 & 1.93 & 0.14 & Region-supported point goal \\
Box + relation offset & Fixed & 66.1 & 29.9 & 0.10 & 0.58 & 1.81 & 0.12 & Free-space relational goal \\
Box + relation + anchor & Fixed & 68.0 & 28.5 & 0.08 & 0.61 & 1.55 & 0.09 & Persistent anchor without uncertainty gate \\
\textbf{Fly0 full} & Fixed & \textbf{70.4} & \textbf{27.2} & \textbf{0.05} & \textbf{0.64} & \textbf{1.24} & \textbf{0.05} & Persistent anchor + uncertainty gate \\
SPF-Qwen + Ego & Adapted & 57.6 & 35.0 & 0.13 & 0.49 & 2.12 & 0.19 & Textual waypoint \\
\bottomrule
\end{tabular}}
\end{table*}

\begin{table}[t]
\caption{Sensor-stack and semantic-update ablations on AerialVLN. NearMiss counts episodes with minimum obstacle clearance below 0.5 m. The pose term denotes the same onboard VIO estimate used by all methods; RTK-GNSS is not part of the online stack.}
\label{tab:ablation}
\centering
\resizebox{\columnwidth}{!}{%
\begin{tabular}{lcccccc}
\toprule
Variant & SR & NE & CR & NearMiss & SPL & Jerk \\
\midrule
\multicolumn{7}{c}{\textit{Sensor stack}} \\
\midrule
RGB only & 48.5 & 43.6 & 0.24 & 0.31 & 0.39 & 2.93 \\
RGB-D & 60.8 & 33.9 & 0.17 & 0.24 & 0.51 & 2.08 \\
RGB-D + uncertainty bridge & 66.2 & 30.6 & 0.11 & 0.18 & 0.57 & 1.74 \\
RGB-D + LiDAR & 68.7 & 28.4 & 0.07 & 0.11 & 0.61 & 1.33 \\
RGB-D + LiDAR + VIO pose & \textbf{70.4} & \textbf{27.2} & \textbf{0.05} & \textbf{0.09} & \textbf{0.64} & \textbf{1.24} \\
\midrule
\multicolumn{7}{c}{\textit{Semantic update frequency}} \\
\midrule
Fly0 @ 0.2 Hz & 65.9 & 31.3 & 0.08 & 0.13 & 0.58 & 1.29 \\
Fly0 @ 0.5 Hz & 70.4 & 27.2 & 0.05 & 0.09 & 0.64 & 1.24 \\
Fly0 @ 1.0 Hz & 71.0 & 26.8 & 0.05 & 0.09 & 0.65 & 1.23 \\
\bottomrule
\end{tabular}}
\end{table}

Table \ref{tab:fairness} addresses the most important fairness concern: whether Fly0 wins only because it uses Ego-Planner. Under a strict matched stack, changing the goal interface from a transient point to a region-supported relational anchor improves SR monotonically from 62.1 to 70.4 while also reducing false-lock rate from 0.16 to 0.05. The comparison to ``SPF-Qwen + Ego'' remains informative, but the first five rows are the key evidence because they hold planner, sensors, pose, MLLM, localization noise, and stopping rules fixed.

Table \ref{tab:ablation} further isolates sensor and update-frequency contributions. Depth and LiDAR both matter, but the uncertainty bridge provides a large gain even before LiDAR is added. Moreover, raising re-grounding from 0.5 Hz to 1.0 Hz offers only a modest improvement, which supports the claim that low-frequency semantic reasoning is sufficient once a stable 3D anchor has been formed.

\subsubsection{Latency and Network Robustness}

\begin{table}[t]
\caption{Latency decomposition of Fly0 on the real platform. Frequencies correspond to nominal operating rates during physical deployment.}
\label{tab:latency}
\centering
\resizebox{\columnwidth}{!}{%
\begin{tabular}{lcccc}
\toprule
Module & Frequency & Mean (ms) & P95 (ms) & Platform \\
\midrule
Image compression & 0.5--1 Hz & 2.3 & 5.1 & Jetson \\
Image uplink & 0.5--1 Hz & 14.8 & 38.2 & Network \\
MLLM inference & 0.5--1 Hz & 612.0 & 931.0 & A800 server \\
JSON parsing + sanity check & 0.5--1 Hz & 2.1 & 4.0 & Jetson \\
ROI depth aggregation & 0.5--1 Hz & 4.7 & 8.5 & Jetson \\
2D-to-3D lifting & 0.5--1 Hz & 1.4 & 2.6 & Jetson \\
LiDAR mapping & 10 Hz & 17.9 & 31.4 & Jetson \\
Ego-Planner update & 50 Hz & 11.6 & 18.7 & Jetson \\
Low-level control & 100 Hz & 3.0 & 4.8 & PX4 \\
\bottomrule
\end{tabular}}
\end{table}

\begin{table}[t]
\caption{Robustness of Fly0-Real performance to communication degradation. Goal age denotes the P95 age of the active anchor at control time.}
\label{tab:network}
\centering
\resizebox{\columnwidth}{!}{%
\begin{tabular}{lcccc}
\toprule
Condition & SR $\uparrow$ & CR $\downarrow$ & Goal age P95 (s) & Fallback \\
\midrule
Normal network & 62.9 & 0.07 & 0.9 & None \\
+100 ms delay & 62.1 & 0.07 & 1.0 & Cached anchor \\
+300 ms delay & 60.4 & 0.09 & 1.6 & Cached anchor \\
+1 s delay & 55.2 & 0.12 & 3.8 & Hover if stale $>$ 4 s \\
5\% packet loss & 58.6 & 0.10 & 2.7 & Re-query + cached anchor \\
Cloud outage (10 s burst) & 53.9 & 0.11 & 4.1 & Hover / return-to-home \\
\bottomrule
\end{tabular}}
\end{table}

Table \ref{tab:latency} clarifies the low-latency claim. The MLLM remains the slowest module, but it is invoked only at 0.5--1 Hz. The onboard geometric loop stays below 20 ms at the 95th percentile, which is compatible with high-frequency flight. Table \ref{tab:network} shows that cached anchors absorb moderate network degradation, but performance degrades once delays exceed the anchor-hold horizon or when cloud outages persist long enough to trigger hover or return-to-home. Across outage episodes, 61\% of collision or near-miss events occur when the active anchor age exceeds 4 s, which is why we present the system as latency-tolerant rather than communication-independent.

\begin{table}[t]
\caption{Anchor reliability and recovery analysis on Fly0-Real. False-lock counts episodes in which an incorrect anchor remains active for more than 3 s.}
\label{tab:anchor}
\centering
\resizebox{\columnwidth}{!}{%
\begin{tabular}{lcccc}
\toprule
Variant & False-lock $\downarrow$ & Recovery $\uparrow$ & SR $\uparrow$ & CR $\downarrow$ \\
\midrule
No jump check & 0.18 & 0.39 & 58.1 & 0.11 \\
+ jump check & 0.11 & 0.55 & 60.3 & 0.09 \\
\textbf{+ multi-view confirm (ours)} & \textbf{0.07} & \textbf{0.68} & \textbf{62.9} & \textbf{0.07} \\
\bottomrule
\end{tabular}}
\end{table}

Table \ref{tab:anchor} examines the main failure mode introduced by persistence itself. Without anchor validation, high-confidence false positives in repeated-object scenes can remain active long enough to create avoidable near-misses. The proposed jump check and multi-view confirmation reduce false-lock rate and improve recovery, which suggests that persistence is useful only when paired with explicit recovery logic.

\subsubsection{Real-World Generalization}
The real-world benchmark is intentionally harder than the average simulation scene: 31\% of episodes contain major occlusion, 28\% contain repeated objects, and 14\% include dynamic obstacles. Under these conditions, Fly0 still achieves 62.9\% SR with 0.07 collision rate. The trajectory examples in Fig. \ref{fig2} and Fig. \ref{fig3}, which are collected in Appendix \ref{apd-trajectories}, show that once the target has been anchored in the world frame, the UAV can continue navigating through temporary visual gaps without repeatedly invoking semantic reasoning at the control rate.

\begin{table*}[!htbp]
\caption{Instruction-type breakdown on Fly0-Real. Shares are measured over the 160 unique instructions, and the final column summarizes the dominant factor behind Fly0's gain over SPF-Qwen.}
\label{tab:breakdown}
\centering
\resizebox{\textwidth}{!}{%
\begin{tabular}{lcccccc}
\toprule
Instruction type & Share & SPF-Qwen SR $\uparrow$ & Fly0 SR $\uparrow$ & Fly0 3D Err. $\downarrow$ & Fly0 CR $\downarrow$ & Main source of gain \\
\midrule
Object only & 22\% & 54.7 & 71.9 & 2.6 & 0.05 & Persistent anchor after first lock \\
Object + direction & 18\% & 46.1 & 66.3 & 3.0 & 0.06 & Better metric target than textual waypoint \\
Object + relation & 24\% & 38.5 & 61.8 & 3.4 & 0.07 & Relation-aware goal offset \\
Multi-step instruction & 12\% & 31.4 & 54.2 & 3.7 & 0.09 & Stable anchor through viewpoint change \\
Ambiguous repeated target & 14\% & 26.8 & 49.5 & 4.2 & 0.10 & Region grounding + re-grounding \\
Target temporarily invisible & 10\% & 23.7 & 52.4 & 3.5 & 0.08 & World-frame goal memory \\
\bottomrule
\end{tabular}}
\end{table*}

\begin{table}[!htbp]
\caption{Per-relation evaluation on the relation-bearing subset of Fly0-Real. SR@2m uses a stricter endpoint tolerance to expose relation errors that may be hidden by the standard 5 m success criterion.}
\label{tab:relation_eval}
\centering
\resizebox{\columnwidth}{!}{%
\begin{tabular}{lcccc}
\toprule
Relation subset & Share & 3D Err. $\downarrow$ & SR@5m $\uparrow$ & SR@2m $\uparrow$ \\
\midrule
front / behind & 18\% & 3.6 & 59.8 & 46.7 \\
left / right & 24\% & 3.3 & 63.1 & 50.9 \\
above / below & 11\% & 2.8 & 66.4 & 54.1 \\
near / around & 21\% & 2.7 & 69.2 & 58.0 \\
nested / compositional & 26\% & 4.5 & 47.6 & 34.2 \\
\bottomrule
\end{tabular}}
\end{table}

\begin{table}[!htbp]
\caption{Trajectory-quality comparison on Fly0-Real under matched planner settings. Lower path ratio, near-miss rate, and control effort indicate more efficient and smoother execution.}
\label{tab:trajquality}
\centering
\resizebox{\columnwidth}{!}{%
\begin{tabular}{lcccccc}
\toprule
Method & SPL & Path Ratio & NearMiss & Replan Hz  \\
\midrule
SPF-Qwen + Ego & 0.42 & 1.43 & 0.19 & 7.6  \\
2D point + depth + Ego & 0.47  & 1.35 & 0.15 & 8.1  \\
\textbf{Fly0 full} & \textbf{0.58} & \textbf{1.21} & \textbf{0.09} & \textbf{8.8} \\
\bottomrule
\end{tabular}}
\end{table}

Table \ref{tab:breakdown} shows that the advantage of Fly0 is not uniform across all instructions. The gain is modest on visually salient object-only commands, but becomes much larger on relation-heavy, ambiguous, and temporarily invisible targets. Table \ref{tab:relation_eval} makes this claim more concrete. The stricter SR@2m metric reveals that nested and front/behind relations remain the hardest categories even when the standard 5 m success threshold is satisfied, which is why we characterize the current relation module as a transparent rule-based approximation rather than a complete language-grounding solution. Table \ref{tab:trajquality} further shows that these gains are accompanied by smoother and shorter trajectories rather than merely more aggressive replanning.

\subsubsection{Model-Scale Sensitivity}

\begin{table*}[!htbp]
\caption{Reference backbone comparison under the fixed Fly0 backend in our April 2026 test environment. Latency is end-to-end semantic-update latency for 896$\times$896 JPEG inputs with temperature 0 and a 128-token JSON budget; cost is provider list price at test time or amortized serving cost per 1,000 queries. These values should be interpreted as environment-specific reference numbers rather than universal model rankings.}
\label{tab:mllm}
\centering
\resizebox{\textwidth}{!}{%
\begin{tabular}{lcccccccc}
\toprule
Model & 2D Acc. $\uparrow$ & 3D Err. $\downarrow$ & SR $\uparrow$ & NE $\downarrow$ & Relation-heavy Acc. $\uparrow$ & Ambiguous Acc. $\uparrow$ & Latency (ms) $\downarrow$ & Cost / 1k $\downarrow$ \\
\midrule
GPT-5 & 87.2 & 2.9 & 71.3 & 26.8 & 88.7 & 84.9 & 1280 & 14.3 \\
Gemini 3 Pro & \textbf{87.6} & \textbf{2.8} & \textbf{71.5} & \textbf{26.6} & \textbf{89.1} & \textbf{85.2} & 1420 & 12.8 \\
Claude 3.7 Sonnet & 86.9 & 3.0 & 70.8 & 27.0 & 87.8 & 84.3 & 1185 & 9.6 \\
\textbf{Qwen2.5VL-32B} & 86.4 & 3.1 & 70.4 & 27.2 & 86.9 & 82.4 & 612 & 1.7 \\
Qwen2.5VL-7B & 82.1 & 3.8 & 67.9 & 29.6 & 81.2 & 76.1 & 318 & 0.4 \\
MiniCPM-V-3B & 79.4 & 4.5 & 65.7 & 31.1 & 77.5 & 72.9 & \textbf{284} & \textbf{0.3} \\
\bottomrule
\end{tabular}}
\end{table*}

Table \ref{tab:mllm} reveals a more nuanced picture than ``all models are the same.'' On simple object-only instructions, all large models exceed 80\% 2D grounding accuracy, so the downstream geometry dominates. The gap appears mainly on relation-heavy and ambiguity-prone commands, where larger models provide a 5--8 point advantage. Qwen2.5VL-32B remains attractive because it is close to the strongest proprietary models in SR and NE while offering much lower latency and cost in this particular setup. These numbers are intended as reference results within our platform, network path, and API configuration rather than as a general ranking across providers.

\begin{table}[!htbp]
\caption{Deployment-budget study on Fly0-Real. Onboard settings remove cloud dependence at the cost of lower semantic refresh rate and reduced image resolution.}
\label{tab:budget}
\centering
\resizebox{\columnwidth}{!}{%
\begin{tabular}{lccccc}
\toprule
Semantic backend & Input px & Uplink KB/frame & Update Hz & SR $\uparrow$ & CR $\downarrow$ \\
\midrule
Cloud Qwen2.5VL-32B & 896 & 86 & 0.5--1.0 & 62.9 & 0.07 \\
Cloud Qwen2.5VL-32B (compressed) & 512 & 41 & 0.5--1.0 & 61.8 & 0.08 \\
Onboard Qwen2.5VL-7B & 448 & 0 & 0.33 & 57.8 & 0.09 \\
Onboard MiniCPM-V-3B & 384 & 0 & 0.25 & 54.1 & 0.10 \\
\bottomrule
\end{tabular}}
\end{table}

Table \ref{tab:budget} clarifies the deployment trade-off created by the cloud semantic backend. Pure onboard operation remains feasible with smaller models, but the lower refresh budget and reduced image resolution degrade grounding quality and downstream task success. Conversely, cloud deployment provides stronger semantics but increases reliance on communication quality. We therefore view cloud inference as a performance-oriented operating point rather than as a prerequisite for the interface itself.

\subsubsection{Failure Taxonomy}

\label{sec:failure_taxonomy}

\begin{table}[!htbp]
\caption{Failure taxonomy observed in real-world deployment. Failures are grouped by dominant source, approximate share, representative scenario, and mitigation strategy.}
\label{tab:failure_taxonomy}
\centering
\small
\setlength{\tabcolsep}{4pt}
\renewcommand{\arraystretch}{1.18}

\begin{tabularx}{\columnwidth}{
@{}
>{\raggedright\arraybackslash}p{0.23\columnwidth}
>{\centering\arraybackslash}p{0.10\columnwidth}
>{\raggedright\arraybackslash}X
>{\raggedright\arraybackslash}X
@{}
}
\toprule
\textbf{Failure type} 
& \textbf{Share} 
& \textbf{Dominant cause / example} 
& \textbf{Mitigation} \\
\midrule

Semantic ambiguity 
& 24\% 
& Repeated or visually similar landmarks, e.g., multiple trees or poles. 
& Clarification prompt and multi-view grounding. \\

Spatial relation error 
& 19\% 
& Misinterpretation of relative phrases such as ``front of the building'' or left/right/above relations. 
& Relation-aware offset estimation with heading consistency check. \\

Depth error 
& 17\% 
& Thin, reflective, or distant targets, e.g., poles, wires, or glass facades. 
& Patch-level depth aggregation and optional LiDAR fusion. \\

Pose drift 
& 12\% 
& Long-horizon transform bias during extended corridor or open-area flight. 
& Periodic VIO refresh and visual-anchor reset. \\

Planner local minima 
& 10\% 
& Dense clutter or narrow passages, e.g., shrubs near a footpath. 
& Global re-planning with corridor or traversability priors. \\

Network delay 
& 9\% 
& Cloud-side grounding unavailable under weak outdoor signal. 
& Cached visual anchors, hover fallback, and return-to-home policy. \\

Dynamic obstacle 
& 9\% 
& Moving pedestrians or bicycles in campus sidewalk scenes. 
& Dynamic keep-out zones and short-horizon motion prediction. \\

\bottomrule
\end{tabularx}
\end{table}

\textbf{Case study: how anchoring succeeds and fails.} In a representative successful Fly0-Real episode, the instruction asks the UAV to approach the front side of a tree beside a footpath and then hover near a low bush region. The first semantic update grounds the tree as an image region and returns the \texttt{front} relation with high confidence. Fly0 aggregates depth over the grounded box, projects the object into the world frame, and synthesizes a visible-side free-space anchor rather than commanding the vehicle to fly toward the tree surface. During the approach, the target is intermittently occluded by foliage and leaves the camera view during a yaw maneuver; however, the LiDAR planner continues to optimize toward the cached world-frame anchor while avoiding shrubs in the local map. When the vehicle reaches the first anchor, the clause parser activates the second sub-goal and re-grounding replaces the old anchor instead of blending two targets. This trace illustrates the intended division of labor: language defines and refreshes the metric goal, whereas high-rate geometry handles smooth and safe execution.

A representative failure has a different structure. In a repeated-object scene, the instruction refers to a tree behind a sign and left of a lamp post, but the semantic branch consistently selects a nearby visually similar tree. The selected instance has valid depth support, and two subsequent views agree geometrically with the same wrong object. As a result, the uncertainty gate and jump check accept a stable but semantically incorrect anchor. The planner then reaches a collision-free endpoint near the wrong tree, producing a navigation failure through large terminal error rather than through unsafe control. This case explains why semantic ambiguity and nested spatial relations dominate Table \ref{tab:failure_taxonomy}: multi-view confirmation can verify anchor stability, but it cannot by itself prove that the correct object instance and relation have been grounded.

The taxonomy therefore separates the remaining errors into three layers. The first is semantic identity and relation grounding, including repeated objects and nested references that require stronger spatial-language disambiguation than the current rule-based interface provides. The second is geometric evidence quality, including unreliable depth, pose drift, and thin or reflective structures that corrupt otherwise correct semantic hypotheses. The third is execution context, including local planner minima, dynamic obstacles, and degraded communication. Once semantic reasoning is decoupled from control, these bottlenecks become explicit and diagnosable rather than being hidden inside a monolithic controller.

\section*{Discussion}
A natural alternative explanation for the gains of Fly0 is \emph{planner bias}: perhaps any method would perform similarly once paired with Ego-Planner. The controlled results in Table \ref{tab:fairness} argue against this interpretation. When the same Qwen2.5VL-32B backbone, the same RGB-D/LiDAR/VIO stack, the same Ego planner, and the same stopping rule are used, performance improves monotonically as the goal interface is strengthened from a transient point to a validated persistent anchor. The remaining gap is not limited to success rate; it also appears in navigation error, collision rate, SPL, jerk, and false-lock rate. This pattern suggests that the observed gain is better explained by the quality of the goal representation than by replacing one controller with another.

A second alternative explanation is \emph{sensor bias}: because Fly0 uses RGB-D, LiDAR, and pose information, one may suspect that the improvement is largely due to additional sensing rather than to the proposed interface. Table \ref{tab:ablation} shows that richer sensing does improve performance, but it also shows that sensing alone is insufficient. Holding the sensor stack fixed, adding the uncertainty-aware bridge on top of RGB-D improves success rate from 60.8 to 66.2 and reduces collision rate from 0.17 to 0.11 before LiDAR is introduced. Likewise, the strict interface-isolation rows of Table \ref{tab:fairness} indicate that access to the same RGB-D, LiDAR, and VIO inputs without persistent goal formation still leaves a substantial gap. These observations suggest that the main contribution should not be interpreted as a particular sensor configuration, but as the mechanism that converts semantic evidence into a world-frame target that can exploit the available sensors effectively.

A third reason the bias explanation is incomplete is that the evidence chain improves upstream of planning. Table \ref{tab:grounding} shows that Fly0 reduces both 2D grounding error and 3D localization error, especially under occlusion. If the gains came primarily from a stronger planner or richer sensors, one would expect most improvements to appear only in downstream motion metrics. Instead, the advantage is already visible at the semantic-grounding stage and persists through projection, safety, and final task completion. This staged pattern is consistent with the claim that Fly0 improves the \emph{interface} between language and control rather than merely the control backend.

Taken together, the controlled comparisons support viewing Fly0 primarily as an interface contribution relative to prior modular grounding-planner systems. The planner enforces local safety, the sensors provide geometric observability, and the semantic-to-geometric interface determines whether language goals can be represented in a form that the planner can use reliably. This interpretation also explains why the gains concentrate on relation-heavy, ambiguous, and temporarily invisible targets in Tables \ref{tab:breakdown} and \ref{tab:relation_eval}: these are precisely the cases in which interface quality, rather than raw planner strength, is most critical.

\section*{Limitations}
Despite the improvements reported here, Fly0 still has several limitations that align with the failure analysis in Sec. \ref{sec:failure_taxonomy}. First, the relation-aware goal generator relies on a compact observation-conditioned rule set rather than a learned spatial-language grounding model, which is why nested and compositional relations remain the weakest category in Table \ref{tab:relation_eval}. The failure case study also shows that geometric consistency is not equivalent to semantic identity: a wrong repeated object can be stable across views and still pass the current anchor checks. Second, the current uncertainty model is local to the grounded region and depth support. Strongly reflective, texture-poor, thin, or distant targets can still degrade 3D goal lifting, and long-horizon pose drift can bias an otherwise correct anchor. Third, the planner is intentionally local and safety-oriented; dense clutter, narrow passages, and moving pedestrians may require global traversability reasoning and dynamic obstacle prediction beyond the present implementation. Fourth, while the real-world benchmark is substantially broader than a simple demonstration set, it remains smaller than long-term autonomy settings involving persistent crowds, severe weather, or GNSS-denied urban canyons. Finally, although the control loop is fully onboard, the strongest semantic results still depend on cloud-hosted MLLMs, which introduces deployment constraints in low-connectivity environments and motivates the smaller onboard alternatives in Table \ref{tab:budget}.


\section{Conclusion}
We presented Fly0 as a persistent metric anchoring framework for zero-shot UAV-VLN. The central idea is to convert open-vocabulary language goals into validated world-frame navigation anchors, allowing low-frequency MLLM reasoning to cooperate with high-frequency geometric control instead of replacing it. To make this interface reliable, we combined structured region grounding, uncertainty-aware 3D lifting, relation-aware navigable goal generation, persistent goal memory, anchor validation, and cloud-edge safety fallback. Across simulation, fairness-controlled comparisons, and real-world flight, this design improves grounding accuracy, navigation success, collision rate, and latency tolerance under the protocols studied in this paper. More broadly, our results suggest that progress in UAV-VLN depends not only on stronger multimodal models, but also on better interfaces that expose uncertainty, preserve actionable geometric structure, and make failure recovery explicit.

\bibliographystyle{IEEEtran}
\bibliography{ref}

\appendices
\onecolumn

\section{Details of Experimental Setup}
\label{appendix a}

To ensure reproducibility and facilitate further research, we provide detailed specifications of our physical aerial platform and the computational infrastructure used in the real-world experiments.

\subsection{UAV Platform and Hardware Configuration}

We utilized a custom-built quadrotor platform, the TTF-400LIVO-RTK, designed for autonomous navigation in complex unstructured environments. The platform is built upon a resilient, collision-tolerant carbon fiber frame with a diagonal wheelbase of 320mm. The core hardware components include:

\textbf{Flight Controller:} A Pixhawk 6C Mini running the PX4 autopilot firmware, equipped with an STM32H743 processor for stable flight attitude control.

\textbf{Onboard Computer:} An NVIDIA Jetson Orin NX (16GB) serves as the edge computing unit. It features an 8-core Arm Cortex-A78AE CPU and a 1024-core NVIDIA Ampere architecture GPU, providing up to 100 TOPS of AI performance. This unit is responsible for sensor data processing, visual transmission, and real-time trajectory generation.

\textbf{Perception Sensors:} 1) \textbf{LiDAR:} A Livox Mid-360 hybrid solid-state LiDAR provides omnidirectional 3D point clouds ($360^{\circ}$ horizontal FOV) at a rate of 200,000 points/sec for obstacle mapping. 2) \textbf{Vision:} An Stereolabs ZED 2i camera is mounted front-facing to capture visual semantics and depth information for the grounding module.

\textbf{Positioning:} For outdoor experiments, we utilized a high-precision RTK GNSS module (OEM-982) with 4G connectivity to provide centimeter-level ground truth for annotation and evaluation only. The online navigation stack consumes PX4/VIO state estimation rather than RTK corrections, so RTK does not act as an additional inference-time sensor for Fly0 or for the fairness-controlled baselines.

\subsection{Computational Architecture and Deployment}

Our system adopts a hybrid Cloud-Edge Collaborative Architecture to balance the computational demands of large-scale semantic reasoning with the real-time requirements of flight control:

\textbf{Server-Side Semantic Reasoning (Cloud):} The computationally intensive Multimodal Large Language Model (MLLM) is deployed on a remote high-performance server cluster equipped with 4 $\times$ NVIDIA A800 GPUs. During operation, the UAV transmits compressed first-person view images to the server through a low-latency API. The server executes the structured semantic grounding module and returns a JSON object containing the target point, target box, relation token, and confidence.

\textbf{Onboard Geometric Planning (Edge):} The NVIDIA Jetson Orin NX onboard the UAV handles all latency-critical tasks locally using ROS Noetic. These tasks include depth aggregation, uncertainty checking, 2D-to-3D lifting, persistent anchor maintenance, LiDAR mapping, Ego-Planner updates, and low-level state monitoring.

\textbf{Control Separation:} Semantic updates run at 0.5--1 Hz depending on uncertainty and target distance, while obstacle mapping runs at 10 Hz, trajectory optimization at 50 Hz, and PX4 low-level control at 100 Hz. This separation is what allows Fly0 to tolerate cloud latency without injecting semantic delays into the flight-control loop.

\subsection{Communication Fallback and Safety Policy}

The communication policy used in all physical experiments is intentionally conservative:
\begin{itemize}[leftmargin=*]
\item If a semantic update is delayed but the last valid 3D anchor is younger than 4 s, the UAV continues tracking that anchor and requests a refresh.
\item If the anchor becomes stale ($>4$ s) and no new valid grounding is available, the UAV switches to hover mode.
\item If the semantic service remains unavailable for more than 10 s, the system executes the predefined return-to-home or assisted-landing routine.
\item If a new semantic update implies an anchor jump larger than $\tau_{jump}$ in a repeated-object scene, the update remains tentative until multi-view confirmation succeeds.
\item Any safety-pilot intervention, emergency stop, or clearance violation below 0.3 m is logged as a failure event during evaluation.
\end{itemize}

\subsection{Real-World Benchmark Protocol}

The Fly0-Real benchmark is designed as a deployment-oriented benchmark rather than a demonstration set. We summarize the protocol below:
\begin{itemize}[leftmargin=*]
\item \textbf{Scene split:} 4 scenes are used for platform shakedown, hardware safety checks, and benchmark construction auditing; 14 scenes are held out for evaluation. The prompt, confidence thresholds, and model parameters are unchanged across these splits.
\item \textbf{Instruction coverage:} The benchmark contains object-centric, relation-heavy, ambiguous, and long-horizon multi-step instructions with repeated objects, partial occlusion, and dynamic obstacles. Instructions are human-authored, then manually paraphrased and audited for referential uniqueness before flight.
\item \textbf{Episode control:} Every method is tested on the same instruction, start pose, and stopping rules. Real flights are repeated 5 times per instruction.
\item \textbf{Failure definition:} An episode fails if the UAV collides, is safety-aborted, times out, or ends farther than 5 m from the target. For relation-bearing subsets, we additionally report SR@2m.
\item \textbf{Ground truth:} Object-only targets are annotated as reachable free-space endpoints near the referent. Relation-bearing targets are annotated by two raters using a canonical object-centered standoff rule and adjudicated if their annotations differ by more than 1 m. Terminal positions are measured with RTK-GNSS for outdoor scenes and cross-checked with survey markers in repeated trials.
\item \textbf{Release scope:} The public repository contains prompts, parsers, benchmark metadata, evaluation scripts, failure tags, and an auditable subset of trajectories and annotations with sensitive imagery and exact GPS removed.
\end{itemize}

\subsection{Baseline Implementation Details}
\label{apd-baselines}

This subsection records how each comparison method is executed or adapted in our evaluation so that the main text can focus on aggregate results.

\textbf{AerialVLN} is evaluated with the official policy and benchmark protocol when available; for Fly0-Real transfer, its action outputs are mapped to the closest feasible UAV motion primitive under the same episode timeout and collision criteria.

\textbf{OpenUAV} uses its released simulator interface and task formulation for simulated evaluation; in controlled comparisons, we keep its language-conditioned policy unchanged and replace only unavailable simulator-only sensing with the matched RGB stream.

\textbf{UAV-Flow} is treated as a supervised flying-on-a-word baseline that maps visual observations and instructions to motion commands; for real-world tests, the command interface is wrapped by the same safety monitor used for all methods, but no Fly0 anchor memory or uncertainty gate is added.

\textbf{OpenFly} is run with its native aerial VLN stack and official evaluation settings on OpenFly-style simulation tasks; when transferred to AerialVLN and Fly0-Real, we preserve its perception-action interface and use the common stopping rule and safety-abort definition.

\textbf{NAVGPT} follows the zero-shot reasoning setting in which the MLLM produces textual navigation decisions from the current observation and task history; we convert these decisions to short-horizon motion primitives and keep the same VIO pose source used by other zero-shot agents.

\textbf{STMR} is implemented as a spatial-representation-enhanced MLLM agent following its published aerial VLN protocol; the planner-facing output remains a text or waypoint decision rather than a persistent uncertainty-gated world-frame anchor.

\textbf{CityNavAgent} is evaluated as a hierarchical semantic-planning baseline with global memory for aerial urban scenes; in cross-benchmark runs, the high-level memory and planning prompts are retained while the low-level execution is constrained by the same safety monitor and episode budget.

\textbf{SPF} is implemented as a learning-free VLM pointing baseline that localizes the goal and generates a flight target from the current view; the ``SPF-Qwen + Ego'' controlled variant uses the same Qwen2.5VL-32B backbone and Ego-Planner backend as Fly0, but keeps SPF's transient waypoint interface without Fly0's relation-aware anchor validation.

\textbf{Interface-isolation variants} are internal controls that progressively add box-supported depth aggregation, relation offsets, persistent anchors, and uncertainty gates while keeping the RGB-D camera, LiDAR map, VIO pose, Qwen2.5VL-32B backbone, Ego-Planner, localization noise, stopping rule, and safety monitor fixed.

\section{Prompt Design for MLLM}
\label{apd-prompt}




The following presents the prompt template used to guide the MLLM in transforming visual inputs and navigation instructions into structured semantic outputs, including target points, target regions, relational markers, and confidence scores. The specific textual format used in the experiments is summarized as follows: all benchmarks and evaluated models adopt the same protocol; differences arise only in the API wrappers provided by individual service providers when JSON mode is supported.

\begin{verbatim}
Task: Given the current UAV image and the active navigation clause, return one
JSON object and no prose.

Schema:
{
  "point": [x, y],
  "box": [x1, y1, x2, y2],
  "relation": "front|behind|left|right|above|below|near|around|inside|between|none",
  "confidence": c
}

Rules:
1. "point" is the representative pixel of the referred target.
2. "box" tightly encloses the referred object or region in the current image.
3. "relation" is the dominant navigational relation in the active clause.
4. "confidence" is a scalar in [0,1].
5. If the target is partially visible, return the best estimate with lower
   confidence.
6. If the active clause refers to a region (e.g., road or grass area), return
   the region box and set "relation" to "none" unless a relation is explicit.
7. Output JSON only.
\end{verbatim}

Malformed JSON triggers one schema-repair retry. The retry prompt includes only the invalid JSON string, the parser error, and the same schema; it does not include any benchmark-specific hint or ground-truth information. Coordinates that fall outside the image by at most two pixels are clipped to the image boundary; larger violations, boxes with width or height below 8 pixels, unsupported relation tokens, non-finite values, or confidence values outside $[0,1]$ cause the semantic update to be rejected and the previous valid anchor to be retained. We do not use benchmark-specific prompt wording beyond inserting the active clause for multi-step instructions.

\section{Implementation Details for Reproducibility}
\label{apd-implementation}

This appendix records the constants used by the semantic-to-geometric interface. The same prompt schema, relation vocabulary, and geometric constants are used across AerialVLN, OpenFly, and Fly0-Real; no benchmark-specific prompt changes or evaluation-scene adaptation is used.

\begin{table}[h]
\caption{Fixed constants used by the Fly0 interface. Confidence thresholds apply to the model-reported confidence $c$ unless otherwise specified.}
\label{tab:impl_constants}
\centering
\renewcommand{\arraystretch}{1.12}
\begin{tabularx}{\textwidth}{p{0.27\textwidth}p{0.18\textwidth}X}
\toprule
Quantity & Value & Role \\
\midrule
Grounding image size & 896$\times$896 & Input resolution for all MLLM semantic updates. \\
Depth support patch & 35\% of the shorter box side, clipped to 21--96 px and rounded to an odd side length & Patch intersected with the grounded box for ROI depth aggregation. \\
Depth range $(d_{min},d_{max})$ & (0.8 m, 35 m) & Valid depth interval for ZED 2i and simulated RGB-D sensors. \\
Valid-depth ratio $\tau_{occ}$ & 0.35 & Rejects regions dominated by missing depth, sky, or heavy occlusion. \\
Depth variance $\tau_{var}$ & 2.25 m$^2$ & Rejects high-discontinuity regions after 10--90 percentile trimming. \\
Depth-confidence scale $\tau_{\sigma}$ & 1.5 m & Converts local depth variance into semantic-geometric confidence. \\
Minimum model confidence & 0.35 & Rejects low-confidence semantic candidates before 3D lifting. \\
Minimum semantic-geometric confidence & 0.25 & Rejects candidates after depth uncertainty has been applied. \\
Fusion range $(\alpha_{min},\alpha_{max})$ & (0.20, 0.80) & Bounds how strongly a single update can move the persistent anchor. \\
Decision boundary and margin & 0.55 $\pm$ 0.10 & Defines the ``close-to-boundary'' interval for tentative updates. \\
Anchor jump $\tau_{jump}$ & 3.0 m & Triggers tentative handling when a candidate moves the anchor abruptly. \\
Agreement radius $\tau_{agree}$ & 1.0 m & Required consistency between consecutive tentative candidates. \\
Semantic update rate & 0.5 Hz / 1.0 Hz & Nominal rate and elevated rate under near-target, high-uncertainty, or repeated-object conditions. \\
Anchor hold / fail time & 4 s / 10 s & Cached-anchor horizon and return-to-home or assisted-landing timeout. \\
Free-space voxel size & 0.2 m & Resolution of the LiDAR ring-buffer occupancy map. \\
Free-space window & 30 m $\times$ 30 m $\times$ 8 m & Local map centered on the UAV for goal projection and collision queries. \\
Obstacle inflation radius & 0.45 m & Clearance inflation used before accepting a free-space goal candidate. \\
Projection lattice and radius & 0.25 m, max 6 m & Expanding search used by $\Pi_{\mathcal{F}_t}$ around the raw relation target. \\
Accepted altitude band & [0.8 m, 8.0 m] & Prevents relation offsets from producing ground strikes or excessive climb. \\
\bottomrule
\end{tabularx}
\end{table}

\begin{table}[h]
\caption{Operational definition of relation offsets. The frame is frozen at grounding time. $\mathbf{e}_{vis}$ points from the object to the UAV, $\mathbf{e}_l$ is the UAV-left axis, and $\mathbf{e}_z$ is the world-up axis. All raw targets are projected to $\mathcal{F}_t$ before execution.}
\label{tab:relation_offsets}
\centering
\renewcommand{\arraystretch}{1.12}
\begin{tabularx}{\textwidth}{p{0.16\textwidth}p{0.26\textwidth}X}
\toprule
Relation & Offset or candidate set & Implementation detail \\
\midrule
\texttt{none} & $\mathbf{0}$ & Uses the lifted object or region centroid directly, followed by free-space projection. \\
\texttt{front} & $+2.0\mathbf{e}_{vis}$ m & Observation-conditioned visible side of the object; avoids assuming object-canonical orientation. \\
\texttt{behind} & $-2.0\mathbf{e}_{vis}$ m & Opposite side from the visible face at grounding time. \\
\texttt{left} & $+2.0\mathbf{e}_l$ m & Left is defined in the frozen UAV heading frame, not the later vehicle heading. \\
\texttt{right} & $-2.0\mathbf{e}_l$ m & Symmetric to \texttt{left}. \\
\texttt{above} & $+1.5\mathbf{e}_z$ m & Applied after object lifting and then clamped to the accepted altitude band. \\
\texttt{below} & $-1.0\mathbf{e}_z$ m & Used only when the projected point remains above the minimum flight altitude. \\
\texttt{near} & $+1.5\mathbf{e}_{vis}$ m & Shorter visible-side standoff than \texttt{front}. \\
\texttt{around} & 8 points on a 2.5 m horizontal ring & Selects the free candidate closest to the incoming direction and with minimum anchor motion. \\
\texttt{inside} & Region centroid & Used for navigable regions such as road, plaza, gate, or courtyard; requires the centroid projection to be free. \\
\texttt{between} & Midpoint of two lifted referents & The clause parser extracts two noun phrases; each is grounded once with the same schema, then their midpoint is projected to free space. \\
\bottomrule
\end{tabularx}
\end{table}

For repeated-object handling, no ground-truth scene labels are used online. The active clause parser extracts the head noun phrase, and the system stores valid and tentative candidates in a 10 s rolling cache keyed by that phrase. A phrase is marked repeated when two candidates for the same phrase are more than 1.5 m apart in 3D or have image-box IoU below 0.3, provided both have model-reported confidence above 0.45. This conservative test is designed to trigger additional confirmation only when the semantic branch is already producing competing anchors; failure to trigger the flag does not create a success condition by itself.

\section{Navigation Trajectory Results}
\label{apd-trajectories}
\begin{figure*}[h]
  \centering
  \includegraphics[width=0.95\textwidth]{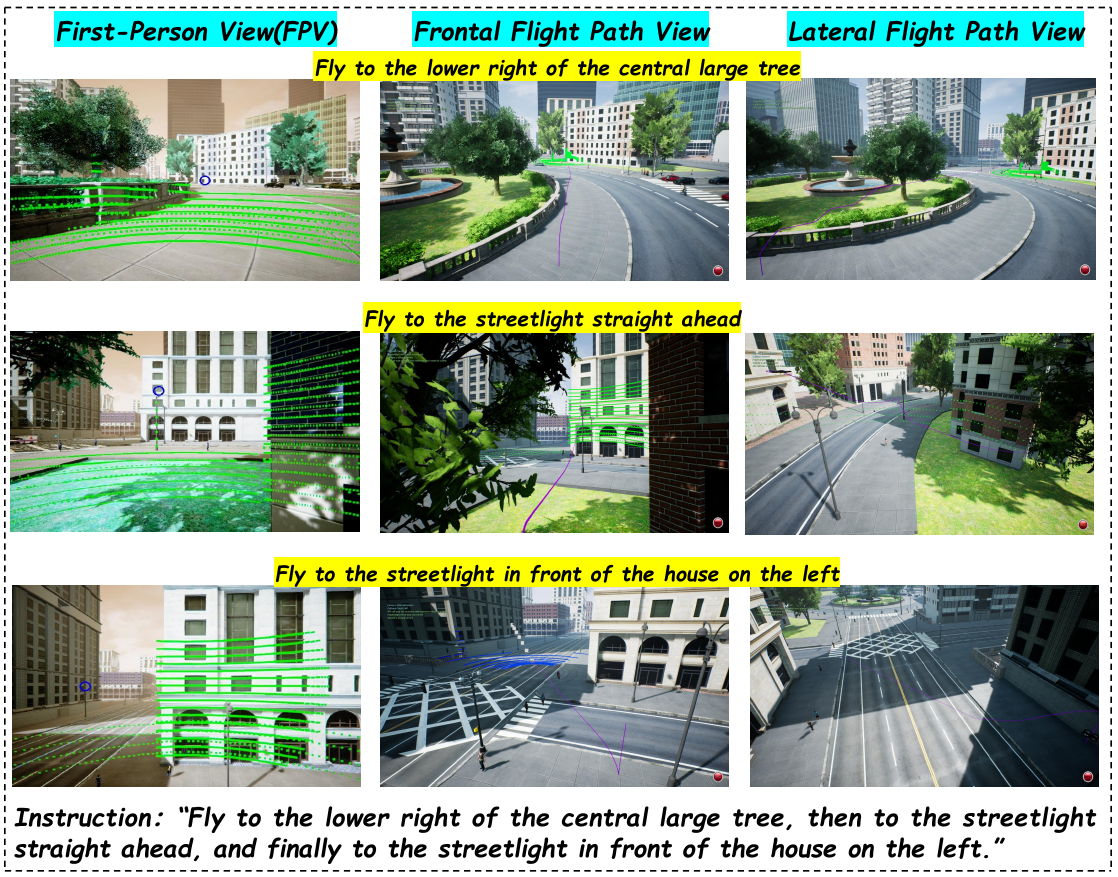}
\caption{Qualitative example of a sequential navigation task in a complex urban environment. The three rows correspond to successive sub-goals, and the columns show the onboard first-person view with LiDAR perception together with frontal and lateral views of the planned trajectory.}
  \label{fig2}
\end{figure*}

\begin{figure*}[h]
  \centering
  \includegraphics[width=0.95\textwidth]{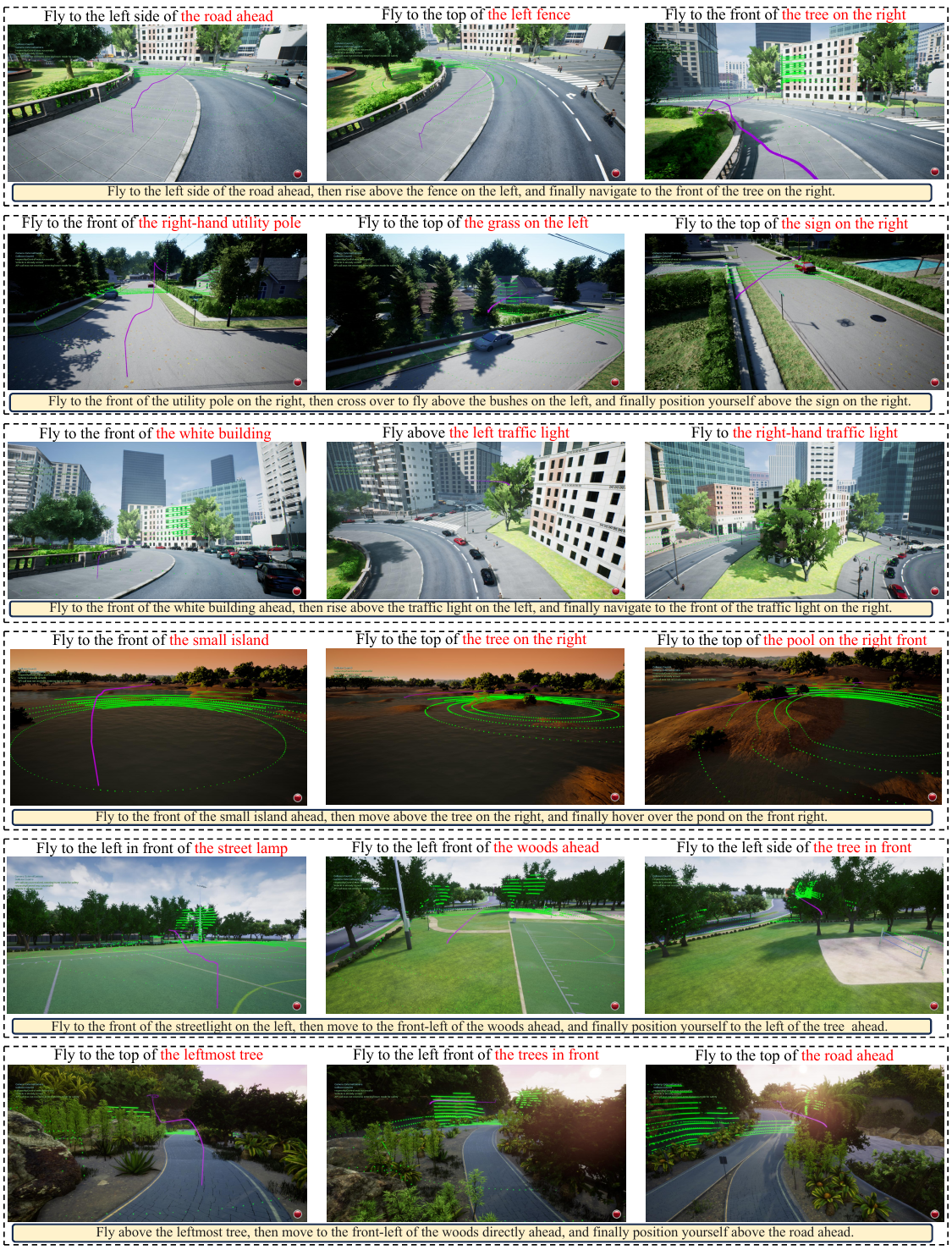}
  \caption{Additional qualitative results in diverse simulation environments. Each row shows an independent long-horizon instruction-following episode in a distinct scene, illustrating the ability of Fly0 to handle heterogeneous topology and fine-grained spatial relations.}
  \label{fig3}
\end{figure*}

Figures \ref{fig2} and \ref{fig3} offer a more comprehensive analysis of the experimental results.

\end{document}